\documentclass[review]{elsarticle}

\usepackage{lineno,hyperref}
\modulolinenumbers[5]
\usepackage{amsmath}
\usepackage{amsfonts}
\usepackage{mathtools}
\usepackage{enumitem}
\usepackage{graphicx}
\usepackage{subcaption}
\usepackage{float}
\usepackage{wrapfig} 
\usepackage{booktabs}
\usepackage{color}
\usepackage[title]{appendix}

\graphicspath{{../images/}}

\newcommand\given [1][]{\:#1\vert\:}

\journal{Medical Image Analysis}

\begin{document}

\begin{frontmatter}

\title{MR Acquisition Invariant Representation Learning}

\author[delft]{Wouter M. Kouw\corref{mycorrespondingauthor}}
\cortext[mycorrespondingauthor]{Corresponding author}
\ead{W.M.Kouw@tudelft.nl}
\author[delft,kope]{Marco Loog}
\author[umcu]{Lambertus W. Bartels}
\author[umcu,nlesc]{Adri{\"e}nne M. Mendrik}

\address[delft]{Delft University of Technology, Mekelweg 4, Delft}
\address[kope]{University of Copenhagen, Universitetsparken 1, Copenhagen}
\address[umcu]{University Medical Center Utrecht, Heidelberglaan 100, Utrecht}
\address[nlesc]{Netherlands eScience Center, Science Park 140, Amsterdam}

\begin{abstract}
Voxelwise classification approaches are popular and effective methods for tissue quantification in brain magnetic resonance imaging (MRI) scans. However, generalization of these approaches is hampered by large differences between sets of MRI scans such as differences in field strength, vendor or acquisition protocols. Due to this acquisition related variation, classifiers trained on data from a specific scanner fail or under-perform when applied to data that was acquired differently. In order to address this lack of generalization, we propose a Siamese neural network ({\sc mrai-net}) to learn a representation that minimizes the between-scanner variation, while maintaining the contrast between brain tissues necessary for brain tissue quantification. The proposed {\sc mrai-net} was evaluated on both simulated and real MRI data. After learning the MR acquisition invariant representation, any supervised classification model that uses feature vectors can be applied. In this paper, we provide a proof of principle, which shows that a linear classifier applied on the {\sc mrai} representation is able to outperform supervised convolutional neural network classifiers for tissue classification when little target training data is available.
\end{abstract}


\end{frontmatter}


\section{Introduction}
Very few of the many medical image analysis algorithms that were proposed in the literature are applicable in clinical practice. One of the reasons for this is the complexity of the medical image data, i.e. the vast amount of variation that is present in this data. A more specific example of this, is brain tissue segmentation in MRI scans. Many automatic methods have been proposed \cite{bezdek1993review,zijdenbos1994brain,clarke1995mri,saeed1998magnetic,pham2000current,suri2002computer,duncan2004geometric,balafar2010review}, but due to a lack of generalization, large scale use in clinical practice remains a challenge \cite{giorgio2013clinical}. In order to test the capacity of algorithms to generalize to new data, a representative sample (dataset) is required. This entails identifying all factors of variation in the data that would influence algorithm performance with respect to the medical image analysis task at hand. For brain tissue segmentation in MRI scans, we identify for example subject related variation (i.e. pathology, age, ethnicity, gender) and acquisition related variation (i.e. MR field strength, protocol settings, scanner vendor, artefacts). Supervised voxel classification approaches have been shown to perform well on small data sets \cite{van2015transfer,moeskops2016automatic,chen2017voxresnet}. However, in order to ensure generalization, these algorithms should be trained and tested on a sufficiently large representative dataset that covers all possible types of variation. This is practically infeasible since training and testing require not only the MRI scans, but also manual labels as ground truth. The manual segmentation process is labor intensive and time consuming, and adds another layer of variation due to non-standardized manual segmentation protocols and inter- and intra-observer variability. To address this problem, we propose an alternative approach, by learning a representation of the data \cite{bengio2013RepLearning} that is invariant to disturbing types of variation, while preserving the variation relevant for the selected classification task, i.e. clinically relevant variation. By reducing undesired variation, this method has the potential to decrease the number of fully labeled samples required for generalization and enable broader use of voxel classification approaches. 

Overcoming acquisition-variation is a relatively new challenge in medical imaging. One particularly interesting approach focuses on weighting classifiers based on how well their training data matches the test data \cite{van2012supervised,van2015transfer,cheplygina2017transfer}. Examples of transfer classifiers include weighted SVM's \cite{van2015transfer} and weighted ensembles \cite{cheplygina2017transfer}. But these methods are very dataset-dependent: the classifiers need to be retrained for every new test dataset. Ideally, we would like to have a method that removes acquisition-variation or extracts acquisition-invariant features. Domain adaptation researchers have proposed representation learning methods that explicitly maximize "domain confusion": if a classifier cannot distinguish between domains then the representation is domain-invariant \cite{tzeng2014deep,ganin2015unsupervised,ganin2016domain}. For MRI scans, different scanners or acquisition protocols constitute different data domains. These representation learning methods are variants of deep neural networks, called domain-adversarial networks. They have two layers in which a loss function is computed: one layer for the task-dependent loss, such as tissue or lesion classification, and one that maximizes domain confusion. The networks learn representations in which the data from each domain overlaps while the different classes become separable \cite{tzeng2014deep}. They are adversarial because the loss layers operate with different objectives, which can make them very difficult to train \cite{ganin2015unsupervised,ganin2016domain,goodfellow2014generative}. A recent paper has applied domain-adversarial networks to segmenting brain lesions \cite{kamnitsas2017unsupervised}. They achieved excellent performance and provided an in-depth analysis of the adversarial training procedure. However, their networks are still very task-dependent: the learned representation works well for brain lesion segmentation but cannot be used for tumor detection for example. It is not a method for learning a general acquisition-invariant representation.

In this paper, we propose to learn a general representation by marking certain factors of variation as desirable and others as undesired \cite{bengio2009learning}. Learning a representation by explicitly minimizing undesirable factors of variation while maintaining desirable factors will produce a task-independent representation, which can be used for a variety of tasks later on. In order to minimize certain factors of variation while maintaining others, we exploit a particular type of neural network, referred to as a Siamese network \cite{bromley1993signature}. Our work was inspired by the work of Hadsell \cite{hadsell2006dimensionality}, who used Siamese neural networks to learn a lighting-invariant representation for airplane images in the NORB \cite{lecun2004NORB} dataset. In this paper we aim to provide a proof of principle for learning an MR-acquisition invariant ({\sc mrai-net}) representation for MR brain tissue segmentation. 

To test {\sc mrai-net} we simulated MRI data (SIMRI \cite{benoit2005simri, aubert2006new, aubert2006twenty}) from a 1.5T scanner and 3T scanner based on acquisition protocols used to acquire real data (Section \ref{sec:data}) and real tissue segmentations from healthy adults (Brainweb). In addition we used real patient data (3T) as provided by MRBrainS \cite{mendrik2015mrbrains}. We acknowledge that the simulated data is idealistic as compared to real patient data. However, experiments in a controlled environment provide a proof of principle to ensure that the method is behaving appropriately. Translation to real patient data is provided by including the MRBrainS data. For the experiments with the simulated data (Section \ref{sec:1sample} and \ref{sec:MultSamples}), the same subject acquired with different acquisition protocols is used. This is however not a prerequisite to train {\sc mrai-net}. For the experiments that use real patient data, different subjects are used. {\sc mrai-net} is not trained by using tissue labels, but with patches labeled as similar or dissimilar. Factors of variation that should be preserved should be labeled as dissimilar, {\sc mrai-net} will aim to reduce all other factors of variation.

\section{Magnetic resonance acquisition-invariant network}\label{method}
Neural networks transform data based on minimizing a loss function. In supervised neural networks, labels are used to determine the loss (error between prediction and label). Many labels are required to learn a task. We aim to use as little labels as possible to learn a representation in which the variation over different methods of acquisition is minimal, without destroying the variation relevant to distinguish between brain tissues. 

The proposed network works as follows. Suppose that we have scans that are acquired in two different ways (A and B). Possible differences can be in field strength, scanner vendor, acquisition protocol, and so on. A tissue patch, for example gray matter, is selected from both scans A and B. The aim is to teach the network that both these patches are gray matter regardless of their acquisition variation. Therefore, we use a loss function that expresses that in the {\sc mrai}  representation, pairs of samples from the same tissue but from different scanners should be as similar as possible. However, that expression alone would cause all samples to be mapped to a single point and would destroy variation between tissues. To balance out the action of pulling pairs marked as similar together, it is necessary to push other pairs apart \cite{hadsell2006dimensionality}. Since we want to maintain the relevant variation between tissues, we additionally express that in the {\sc mrai} representation, pairs from different tissues should retain their dissimilarity. The loss function is described in section \ref{sec:siam_loss}. Section \ref{sec:sim_labeling} describes how pairs of samples are labeled as similar or dissimilar. The  Siamese neural network that is used to learn the {\sc mrai} representation is described in section \ref{sec:net_arch}. The network consists of two pipelines with shared weights and a Siamese loss layer that acts on the output layer of the two pipelines ({\sc mrai} representation).  

\subsection{Siamese loss} \label{sec:siam_loss}
Neural networks transform data in each layer. We summarize the total transformation from input to output with the symbol $f$, i.e. patch $a$ will be mapped to the new representation with $f(a)$ and patch $b$ will be mapped with $f(b)$. To find an optimal transformation, we employ a loss function based on distances between pairs of patches in the output representation, i.e. $\| f(a) - f(b)\|$. Pairwise distances are computed through an $L^1$-norm, denoted by $\| \cdot \|_1$. We used an $L^1$-norm as opposed to for instance an $L^{2}$-norm, because larger values of $p$ in $L^p$-norms either result in problems in high-dimensional spaces or result in problems with the gradient during optimization (see Appendix \ref{app:norms}). 

The loss function for the similar pairs consists of the squared distance, $\ell_{\text{sim}}(f \given a,b) = \left(\| f(a) - f(b) \|_1 \right)^2$. We chose this formulation in order to express that large distances are less desirable. The loss function for the dissimilar pairs consists of a hinge loss, where the distance is subtracted from a margin parameter $m$ and the negative values are set to $0$: $\ell_{\text{dis}}( f \given a,b) = \max(0,m-\|f(a) - f(b) \|_1)$. Pairs that lie close together will suffer a loss, while pairs that are pushed sufficiently apart, i.e. past the margin, will not suffer a loss. We discuss the effect of the margin parameter in Section \ref{exp:margin}. 

Each pair of patches is marked with a similarity variable; $y=1$ for similar and $y=0$ for dissimilar. Using the similarity label we can combine the similar and dissimilar loss functions into a single loss function:
\begin{align}
	\ell(f \given {\cal D}) =& \sum_{i} \ y_i \ \ell_{\text{sim}}\left(f \given a_i,b_i \right) + (1-y_i) \ \ell_{\text{dis}} \left( f \given a_i,b_i) \right) \nonumber \\
	=& \sum_{i} \ y_i \| f(a_i) - f(b_i) \|_1^2 + (1-y_i) \max\left(0, m - \| f(a_i) - f(b_i) \|_1 \right) \, . \nonumber
\end{align}
where $i$ iterates over pairs and ${\cal D}$ refers to the whole dataset of pairs.

This type of loss function is known as a \emph{Siamese} loss \cite{bromley1993signature,hadsell2006dimensionality}. Note that it is asymmetric: it penalizes samples from one class differently than samples from another class.

\subsection{Labeling pairs as similar or dissimilar} \label{sec:sim_labeling}
\begin{figure}[h!]
\centering
\includegraphics[width=.9\textwidth]{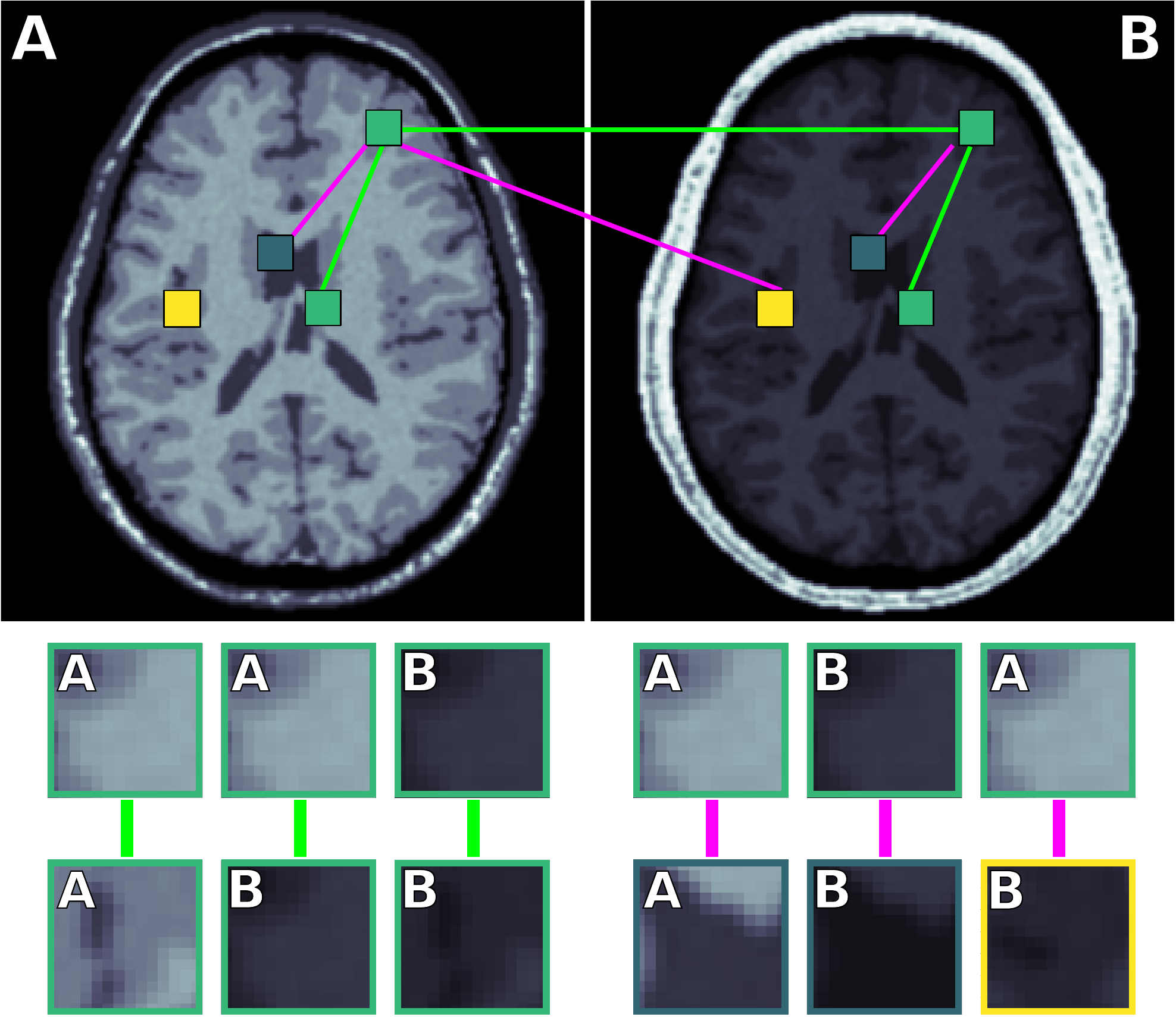}
\caption{Illustration of extracting pairs of patches from images from scanner A and B. (Top) Each image shows 4 patches: 2 gray matter ones (green), 1 cerebrospinal fluid (blue) and 1 white matter (yellow). The lines mark the 6 types of combinations from Section \ref{sec:sim_labeling}. Green lines indicate similar pairs and purple lines indicate dissimilar pairs. (Bottom) Enlarged patches belonging to the 6 pairs marked in the top images.
}
\label{fig:sim_labeling}
\end{figure}

As described above, suppose we have two medical images from two different scanners; A and B. Assume that we have sufficient manual segmentations (labeled voxels) on scans from scanner A, to train a supervised classifier, but a very limited amount of labels from scanner B, for example $1$ labeled voxel per tissue for $1$ subject. The data from scanner A will be referred to as the source set, and the data from scanner B as the target set. Let $K$ be the set of tissue labels. The set of patches extracted from Scanner A is denoted $\{(a_{tn}\}_{n=1}^{N}$, and the set from scanner B is denoted $\{b_{tm}\}_{m=1}^{M}$, with $t$ specifying the sample's tissue. Given these two sets of patches, we form sets of similar and dissimilar pairs, with a similarity label $y$. The following pairs are labeled as similar ($y=1$) and therefore will be pulled closer together: 
\begin{itemize}
\item Source patches from the same tissue $k \in K$: $\{(a_{t=k},a_{t=k})\}$,
\item Source and target patches from the same tissue $k \in K$: $\{(a_{t=k},b_{t=k})\}$,
\item Target patches from the same tissue $k \in K$: $\{(b_{t=k},b_{t=k})\}$.
\end{itemize}
The subscript $t=k$ selects all patches that belong to tissue $k$. The following pairs are labeled as dissimilar ($y=0$) and therefore will be pushed apart: 
\begin{itemize}
\item Source patches from different tissues $k,l \in K$: $\{(a_{t=k},a_{t=l})\}$,
\item Source and target patches from different tissues $k,l \in K$: $\{(a_{t=k},b_{t=l})\}$,
\item Target patches from different tissues $k,l \in K$: $\{(b_{t=k},b_{t=l})\}$.
\end{itemize}
Figure \ref{fig:sim_labeling} illustrates the process of selecting pairs of patches from different scanners. Consider a medical image from scanner A and scanner B, with 2 GM patches (green), 1 WM patch (yellow) and 1 CSF patch (blue) for each image. Using these patches we can generate the following pairs: a GM patch from A with another GM patch from A $(a_{t=k}, a_{t=k})$, a GM patch from A with a GM patch from B $(a_{t=k}, b_{t=k})$, a GM patch from B with another GM patch from B $(b_{t=k}, b_{t=k})$, a GM patch from A with a CSF patch from A $(a_{t=k}, a_{t=l})$, a GM patch from B with a WM patch from B $(a_{t=k}, b_{t=l})$, and a GM patch from B with a CSF patch from B $(b_{t=k}, b_{t=l})$. The bottom of the image shows examples of these 6 pairs of patches.

The pairs are concatenated into a dataset ${\cal D} = \{(a_i,b_i,y_i)\}_{i=1}^C$, where $i$ iterates over the pairs. In total, the number of combinations is $C = \sum_{k \in K} (N_k + M_k)^2 + \sum_{(k,l) \in {K \choose 2}} (N_k N_{l} + N_{k} M_{l} + M_{k} M_{l})$, where  $N_k$ refers to the number of source patches from the $k$-th tissue and, likewise, $M_k$ refers to the number of target patches from the $k$-th tissue. The number of pairs that can be generated is very large, even when only a small number of patches is available. For example, taking $10$ patches of 3 tissues from 4 source scans and 1 patch of 3 tissues from 1 target scan, results in $2784$ pairs of patches that can be used for training the deep neural network.

\subsection{Network architecture} \label{sec:net_arch}
\begin{figure}[h!]
\centering
\includegraphics[width=1.0\textwidth]{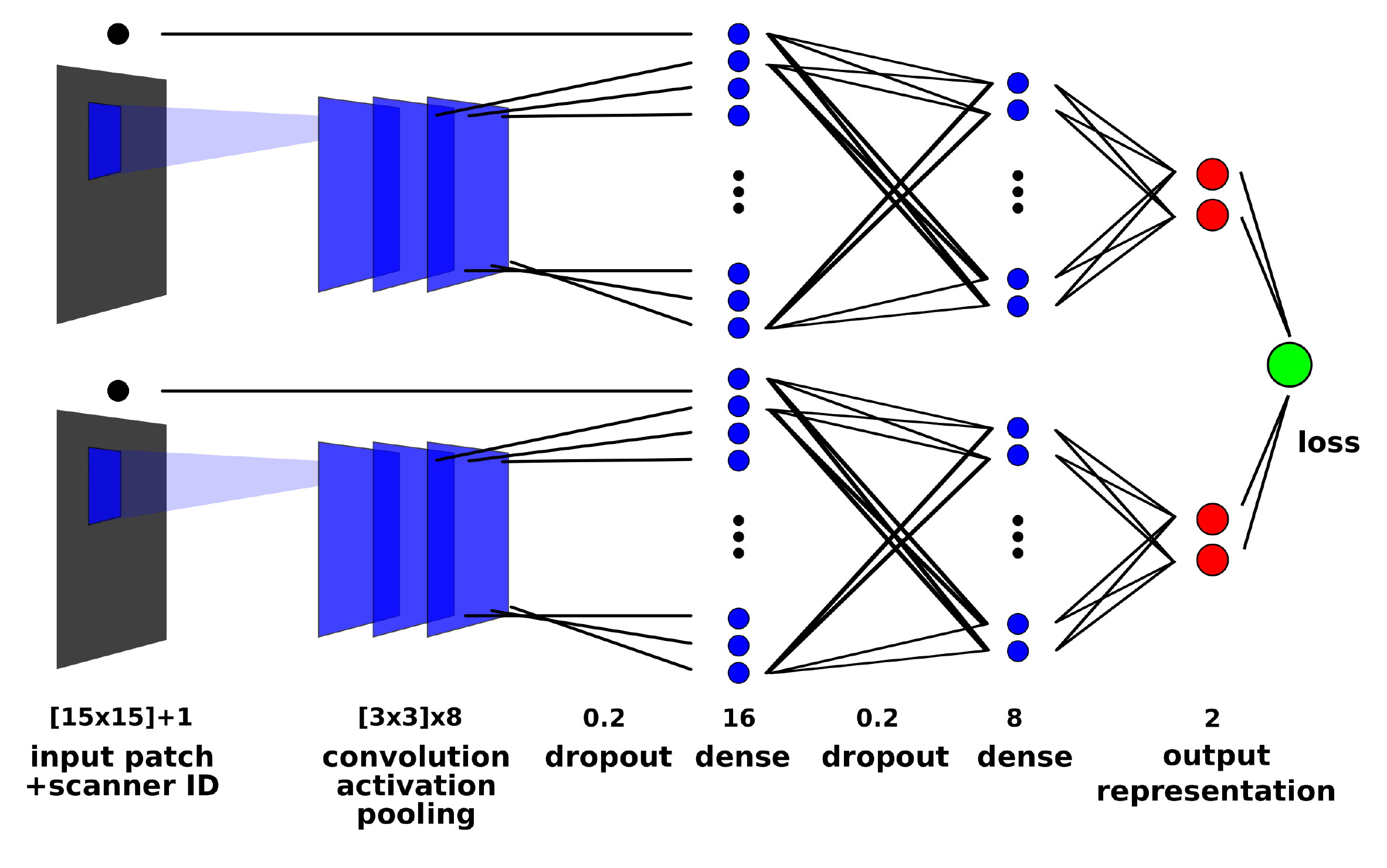}
\caption{Schematic of {\sc mrai-net}'s architecture. Pairs of patches are fed into two pipelines that share parameters (i.e. produce the same mapping). The red nodes depict the representation in the final layer, while the green node depicts the loss function.}
\label{fig:net_arch}
\end{figure}

Figure \ref{fig:net_arch} shows a diagram of the network architecture. The network consists of two pipelines and a Siamese loss layer that acts on the output layers (red nodes). Pairs of patches enter the input layer (black squares) where they are convolved (blue squares) and mapped to feature vectors (blue nodes). The final layer is a low-dimensional feature space (red nodes). The Siamese loss layer (section \ref{sec:siam_loss}) calculates the distance between each pair in their new representation and computes the loss based on whether the pair is marked as similar or dissimilar. The two pipelines share their weights, which means they are constrained to perform the same transformation. During training, the loss is propagated back through the network, adjusting the network weights. 

Width and depth of the network may vary. In this paper, we made the following choices: input patches are size $[15 \times 15]$ and scanner identification is set to a single variable. The convolution block consists of $8$ kernels of size $[3\times 3]$ with a rectifying linear unit (ReLU) activation function and a max-pooling layer of size $[2 \times 2]$. The output of these operations is flattened and the scanner ID ($0$ for source, $1$ for target) is appended. The scanner ID ensures that regions of different tissues in different scanners do not overlap in the input space. The flattened and pooled convolutional layer output, plus the scanner ID is then densely mapped to a 16-dimensional representation. A dropout noise of size $0.2$ is set for each edge. This $16$-dimensional representation is then densely mapped, again with a dropout of $0.2$, to an $8$-dimensional representation, which is finally mapped to a $2$-dimensional representation. We chose a final representation of 2 dimensions because this allows for scatter plot visualizations.

Our method is implemented in a combination of Tensorflow\footnote{\url{https://www.tensorflow.org/}} and Keras\footnote{\url{https://keras.io/}} \cite{45381,chollet2015keras}. This proof of principle uses a 4-layer hybrid convolutional-dense network for the pipeline. However, the network architecture can be changed. Variations involve, for example, more layers, wider layers, larger convolution kernels, and heavier max-pooling. See Section \ref{sec:exp-numparams} for an experiment that varies the layer widths in the network.

\subsubsection{Regularization} \label{sec:reg}
During training, we apply an $l_2$-regularization of $0.001$ to every layer with weights. Regularization punishes the size of the weights, which prevents model over-complexity. In our experiments, the regularization parameter could be increased or decreased by two orders of magnitude with little effect on the networks performance. It is however always necessary to include \emph{some} regularization as there is not only the danger of overfitting to training data but also the danger of overfitting to the specific target subject used for training.

\subsubsection{Optimization}
All experiments in this paper are performed with the default backpropagation algorithm "RMSprop", which normalizes the gradient update with a running average of itself \cite{bottou2012stochastic}. Its default parameters are: a learning rate of 0.001, a $\rho$ of 0.9, an $\epsilon$ of 1e-08, and a weight decay factor of 0.0 (see \cite{bottou2012stochastic} for more details on optimizer parameters). RMSprop is based on stochastic gradient descent, which splits the dataset into batches and updates the networks parameters after processing each batch. An epoch is the number of times the optimization procedure splits the training set into batches. The number of epochs cannot be too large, otherwise the network starts to overfit to the specific target subject from which the target patches originated.

During experimentation we found that it is important that the batches are well-mixed with respect to the 6 types of pairs outlined in Section \ref{sec:sim_labeling}. If this is not the case, such as when one batch mostly consists of similar gray-matter patches and another batch consists mostly of dissimilar gray-matter / white-matter patches, then the network tends to push and pull in the same direction. These actions cancel each other out. The overall effect of having too many uniform batches is that the optimization procedure is slowed down.

\subsection{Training and applying {\sc mrai-net} for adaptive segmentation}
\begin{figure}[h!]
\centering
\begin{subfigure}[t]{.48\textwidth}
	\includegraphics[width=.98\textwidth]{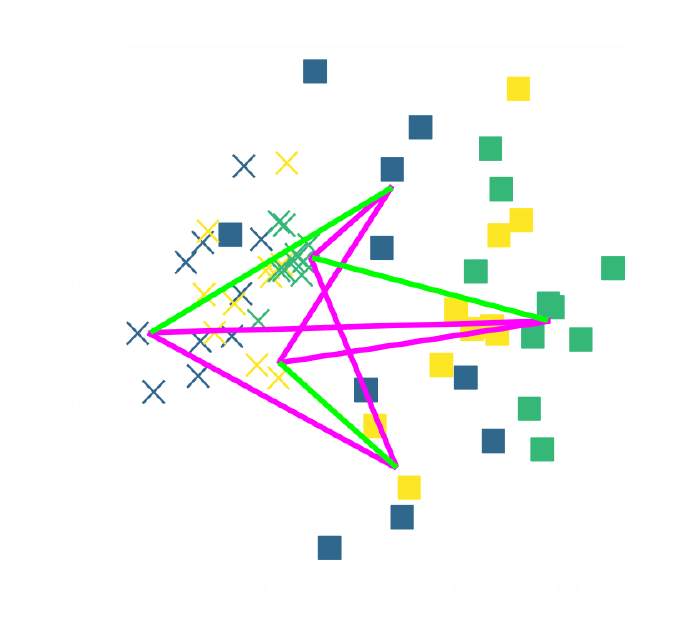}
    \caption{Representation before training.}
    \label{fig:train_net:before}
\end{subfigure} \
\begin{subfigure}[t]{.48\textwidth}
	\includegraphics[width=.98\textwidth]{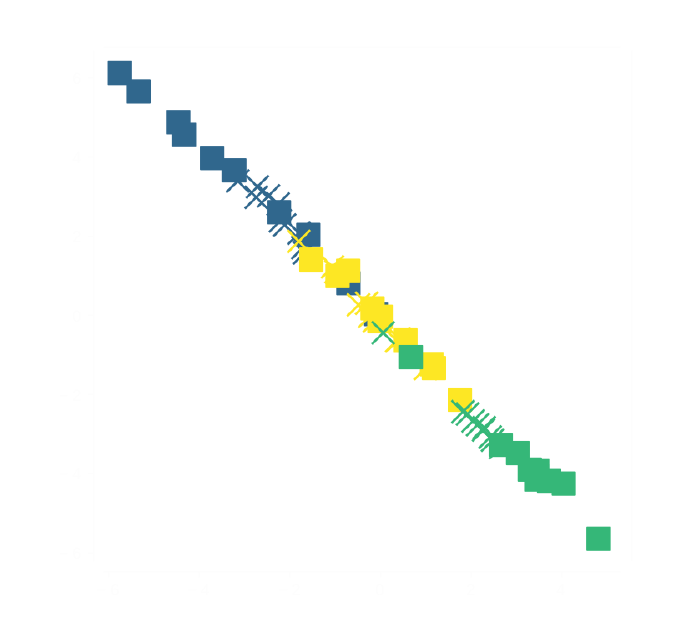}
    \caption{Representation after training.}
    \label{fig:train_net:after}
\end{subfigure}
\caption{Conceptual visualization of {\sc mrai-net}'s training procedure: the network pulls the similar pairs (green lines) closer together and pushes dissimilar pairs (purple lines) apart until it learns a representation in which the variation between scanners is minimal while the variation between tissues is maintained.}
\label{fig:train_net}
\end{figure}

\begin{figure}[h!]
\centering
\includegraphics[width=.95\textwidth]{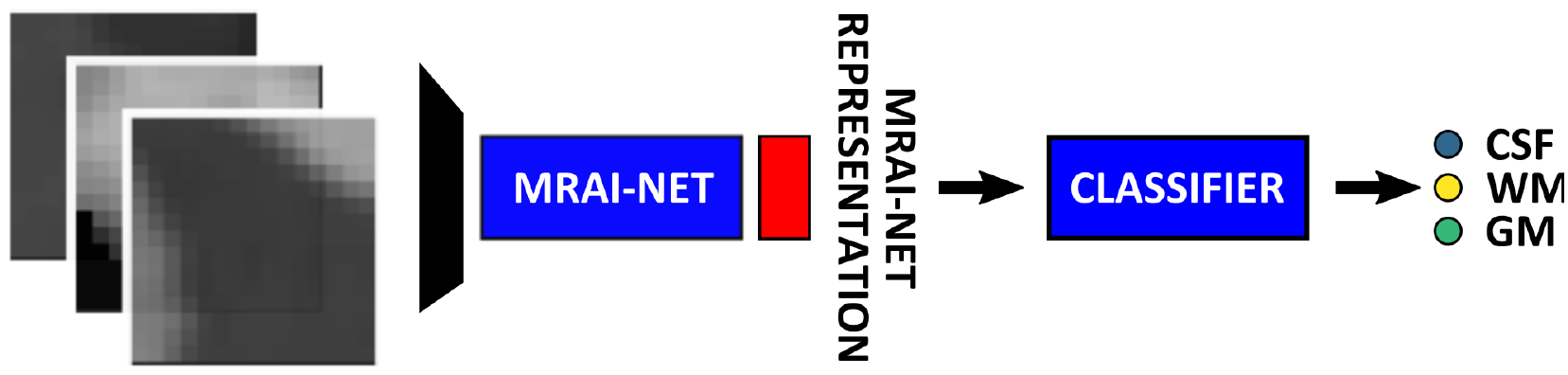}
\caption{A dataset of tissue-labeled single patches is fed through {\sc mrai-net} and represented in the acquisition-invariant space. Subsequently, a classifier is trained to distinguish tissues. A new image is decomposed into patches and fed through the network as well. The trained tissue classifier then makes a prediction for each patch. The predictions are then reshaped back into an image, resulting in the tissue segmentation.}
\label{fig:train_in_net}
\end{figure}

Figure \ref{fig:train_net} illustrates the training procedure of {\sc mrai-net}. Once it is trained and an MR acquisition-invariant representation is learned, it can be used as a preprocessing step for tissue segmentation (Figure \ref{fig:train_in_net}). Because of the shared weights, either one of the pipelines can be used to transform the input patches into the {\sc mrai} representation. Input patches from both the source and target scanner can be fed into the network, and any supervised classification model that uses feature vectors can subsequently be trained to distinguish tissues in the acquisition-invariant representation. Once the supervised classifier is trained, both the trained {\sc mrai-net} and the trained supervised classifier are used to segment a new image. This is done by feeding a new patch through the {\sc mrai-net} and letting the tissue classifier predict the label in the MR acquisition invariant space. In this way, the {\sc mrai-net} acts as a preprocessing step to ensure that acquisition-based variation does not affect the tissue classifier.

\section{Evaluating {\sc mrai-net}}
Since the aim of the {\sc mrai-net} is to preserve variation between tissues while reducing the MR acquisition related variation, two different measures of performance are used to evaluate {\sc mrai-net}. MR acquisition invariance is measured with the proxy ${\cal A}$-distance that measures the distance between the source and target scanner patches, as described in section \ref{sec:MRAImeasure}. The preservation of tissue variation is measured using the tissue classification performance, and compared to supervised classification with CNN (Section \ref{sec:TissuePreservation}). Section \ref{sec:data} describes the simulated (Brainweb 1.5T, Brainweb3.0T) and real data (MRBrainS) used for the experiments. For each experiment a source and target domain was specified. Four source subjects (100 random patches per tissue) and 1 target subject (1-1000 patches per tissue depending on experiment) were used for training. Four independent target subjects (100 random patches per tissue) were used for testing. Four experiments were set-up: 1) Only 1 patch per tissue from the target domain subject is used for training both the supervised CNNs ({\sc source}, {\sc target}) as well as the {\sc mrai-net} followed by a linear classifier on the simulated data (Brainweb1.5T, Brainweb3.0T), 2) Multiple target training samples per tissue (randomly selected with 50 repeats) are used for training the {\sc source}, {\sc target}, and {\sc mrai	-net} classifiers for both simulated (Brainweb3.0T) and real patient data (MRBrainS). The first experiment (Section \ref{sec:1sample}) was set-up to test if only 1 target patch per tissue would be sufficient in order to learn an MR-acquisition invariant representation. If so, then calibrating a supervised segmentation algorithm for a new scanner using {\sc mrai-net} would require only three clicks in one scan acquired with a new scanner. The second experiment (Section \ref{sec:MultSamples}) illustrates the performance of the {\sc mrai-net} compared to the {\sc target}, and {\sc mrai-net} classifiers when adding more target training samples (Figure \ref{fig:results_b1b3}). Results of using 1 patch per tissue and 100 patches per tissue from the target subject for training are shown in Figures \ref{fig:preds_b1b3}-\ref{fig:preds_b3mb}. The third experiment (Section \ref{sec:exp-numparams}) looks at the performance of the network if we vary the number of convolution kernels and the number of nodes in the dense layers. For the setting where Brainweb1.5T is the source scanner and Brainweb3.0T is the target scanner, the network will keep gaining in performance at the cost of adding tens of thousands more parameters. Finally, the fourth experiment (Section \ref{exp:margin}) shows the influence of the margin parameter on the Siamese loss function. If the margin parameter is set too low, tissue variation will not be preserved. On the other hand, if the margin parameter is set too high the acquisition variation will not be reduced. The next two sections describe how these two types of variation are measured.

\subsection{MR acquisition invariance measure}\label{sec:MRAImeasure}
The ${\cal H}$-divergence can be used as a measure of the discrepancy between the source and target scanner data sets \cite{kifer2004detecting,ben2007analysis,ben2010theory}. This divergence relies on the ability of a classifier to distinguish between domains. If a classifier is not able to distinguish source from target, i.e. has a test error of $1/2$, then invariance is achieved. Unfortunately, the original ${\cal H}$-divergence is a measure between distributions and not samples. Since we only have samples, we use its proxy instead: the ${\cal A}$-distance \cite{ben2007analysis,ben2010theory}, as used in \cite{ganin2016domain}. The proxy ${\cal A}$-distance, denoted by $d_{\cal A}$, is defined as follows:
\begin{align}
	d_{\cal A}(x,z) = 2(1 - 2 e(x,z)) \, , \label{eq:proxyA}
\end{align}
where $e$ represents the test error of a classifier trained to discriminate source samples $x$ from target samples $z$. If the source and target data lie far apart, the error will be close to $0$, i.e. perfect separability, and the proxy ${\cal A}$-distance will be close to $2$. If the source and target data overlap, the error will be around $0.5$, i.e. no separability (invariance), and the proxy ${\cal A}$-distance will approach $0$. We use a linear support vector machine (SVM) as domain classifier. 

\subsection{Measure of preserving tissue variation}\label{sec:TissuePreservation}
The tissue classification error is used as a measure of tissue variation preservation. The aim is to learn a linearly separable representation with {\sc mrai-net}, to aid the number of methods that can be used for classification. Therefore, we evaluate the tissue classification error of the samples in the acquisition-invariant representation with a logistic regressor. The classifier is $\ell_2$-regularized and cross-validated for optimal regularization parameters. This classifier {\sc mrai-net}, based on the {\sc mrai-net}, is compared to two other supervised classifiers: 1) {\sc source} classifier: a convolutional-dense neural network (CNN) trained on samples from the source (4 subjects) and target data (1 subject), and 2) {\sc target} classifier: a CNN trained on samples from the target data (1 subject). In order to ensure that differences in performance between {\sc source}, {\sc mrai-net} and {\sc target} are not due to differences between classifiers, the {\sc mrai-net} (Figure \ref{fig:net_arch}) neural network architecture was used for the {\sc source} and {\sc target} classifiers as well. 

\subsection{Data}\label{sec:data}
\begin{figure}[h!]
\centering
\begin{subfigure}[b]{112pt}
	\includegraphics[width=.98\textwidth]{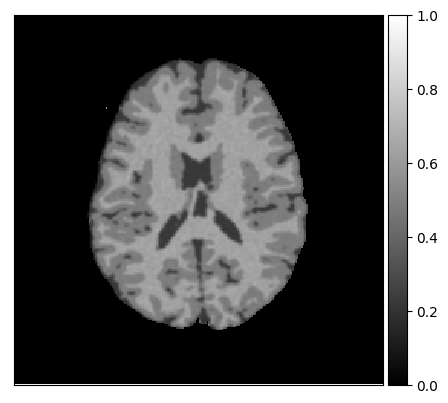}
 	\caption{Brainweb1.5T}
    \label{fig:data:b1}
\end{subfigure} \hfill
\begin{subfigure}[b]{112pt}
	\includegraphics[width=.98\textwidth]{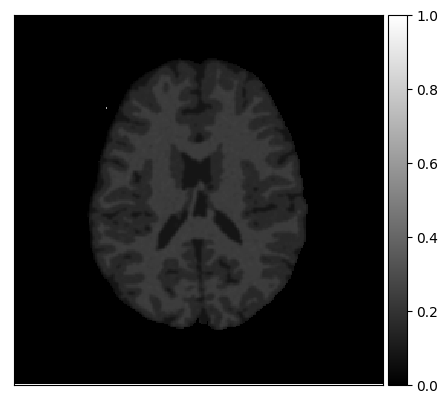}
 	\caption{Brainweb3.0T}
    \label{fig:data:b3}
\end{subfigure} \hfill	
\begin{subfigure}[b]{112pt}
	\includegraphics[width=.98\textwidth]{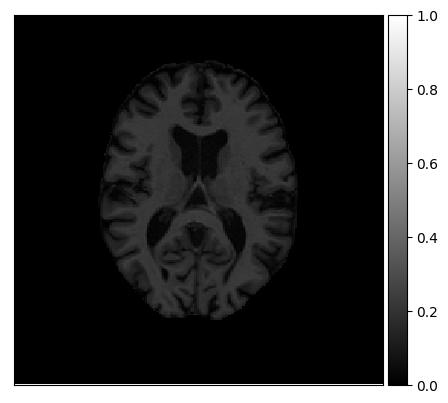}
 	\caption{MRBrains}
    \label{fig:data:mb}
\end{subfigure}
\caption{Example of an MRI scan of a Brainweb anatomical model simulated with SIMRI with a 1.5T protocol (a) and a 3.0T protocol (b), and a real patient scan (MRBrainS) acquired with a 3.0T protocol (c).}
\label{fig:data}
\end{figure}

To be able to provide a proof of principle, we simulated different MR acquisitions from various anatomical models of the human brain \cite{collins1998design,aubert2006twenty}, using an MRI simulator (SIMRI \cite{benoit2005simri, aubert2006new, aubert2006twenty}). The anatomical models consist of transverse slices of 20 normal brains and are publicly available through Brainweb\footnote{\url{http://www.bic.mni.mcgill.ca/brainweb/}}. These models were used as input for the MRI simulator. For the experiments, we simulated two acquisition types: $1)$ Brainweb1.5T, a standard gradient-echo acquisition protocol for a 1.5 Tesla scanner (c.f. \cite{ikram2015rotterdam}), and $2)$ Brainweb3.0T, a standard gradient-echo protocol for a 3.0 Tesla scanner (c.f. \cite{mendrik2015mrbrains}). Table \ref{tab:acqparams} describes the parameters used for the simulation: magnetic field strength (B0), flip angle ($\theta$), repetition time (TR), echo time (TE). Magnetic field inhomogeneities and voxel inhomogeneity (partial volume effects) were not included in the simulation.
\begin{table}[h!]
\centering
\caption{SIMRI Acquisition parameters for the simulation of the Brainweb1.5T and Brainweb3.0T data sets.}
\label{tab:acqparams}
\setlength\tabcolsep{15pt}
\begin{tabular}{ l | c c c c}
		& B0 & $\theta$ & TR & TE \\
\hline 
 Brainweb1.5T & 1.5 Tesla & $20^{\circ}$ & 13.8 ms & 2.8 ms\\ 
 Brainweb3.0T & 3.0 Tesla & $90^{\circ}$ &  7.9 ms & 4.5 ms
\end{tabular}
\end{table}
Appendix \ref{app:NMR} describes the nuclear magnetic resonance (NMR) relaxation times for the tissues in the Brainweb anatomical models, for 1.5 and 3.0 Tesla field strengths. The tissues in the anatomical models are grouped into "background" (BKG), "cerebrospinal fluid" (CSF), "gray matter" (GM), and "white matter" (WM) to compose the ground truth segmentation labels for the simulated scans. The simulations result in images of 256 by 256 pixels, with a 1.0x1.0mm resolution. Figures \ref{fig:data:b1} and \ref{fig:data:b3} show examples of the Brainweb1.5T and Brainweb3.0T scan of the same subject. For all scans, we used a brain mask to strip the skull.
In order to test the proposed method on real data, we use the publicly available training data (5 subjects) from the MRBrainS challenge\footnote{\url{http://mrbrains13.isi.uu.nl/Figure}}. The acquisition parameters used for simulating the Brainweb3.0T are based on the MRBrainS acquisition protocol (3.0T scanner, gradient-echo, B0 = 3.0T, $\theta$ = $90^{\circ}$ flip angle, TE = 4.5ms, and TR = 7.9ms). Figure \ref{fig:data:mb} shows an example of an MRBrainS scan. Again, a brain mask is used to strip the skull. 

\subsection{Experiment 1: One training target sample per tissue} \label{sec:1sample}
The first experiment with the simulated data tests the scenario described at the beginning of this section: suppose a supervised classification algorithm trained on one scanner needs to be calibrated for a new scanner, would this be possible with three clicks (1 for each tissue type) using {\sc mrai-net}? To study this, we manually selected 1 patch for each tissue in the target scan (1 subject) and used this data to train {\sc mrai-net}. Once {\sc mrai-net} has been trained and an acquisition-invariant representation has been learned, we compute the proxy ${\cal A}$-distance and perform a tissue classification experiment.

For computing the proxy ${\cal A}$-distance, we used scans from 10 source subjects and 10 target subjects that had been held back (i.e. we did not draw samples from them to either train {\sc mrai-net} or train any of the tissue classifiers). We randomly drew 50 patches per tissue from each subject, resulting in two sets of 1500 patches. These patches were fed into {\sc mrai-net} which mapped them to the new acquisition-invariant representation. The datasets were labeled $0$ and $1$ for source and target. Next, we trained a linear classifier with 5-fold cross-validation to obtain a test error on data set discrimination. Finally, using this test error and Equation \ref{eq:proxyA}, we computed the proxy ${\cal A}$-distance.

For evaluating the tissue classification performance, we used scans from 10 target subjects that had been held back. From these 10 scans, we drew 50 patches per tissue at random, for a total of 1500 patches. We computed the error rate by computing the proportion of wrong predictions on this test set. We trained the following three classifiers (described in Section \ref{sec:TissuePreservation}): firstly, the {\sc source} classifier (CNN) was trained on images from the source dataset, and applied to the test set to make predictions. Secondly, we trained a linear classifier on the source data mapped to {\sc mrai-net}'s representation. We mapped the test data to {\sc mrai-net}'s representation as well and applied the trained linear classifier to make predictions. Its performance on the test set is indicated with {\sc mrai-net} in Table \ref{tab:b1b3}. The {\sc target} classifier (CNN) was applied to the available target patches. In this experiment, there were 3 target patches in total, which is far too little data to train such a large convolutional network. We included its performance to indicate that using the {\sc target} classifier in this kind of situation is not a sensible option.

For comparison, we performed the same experiment but with randomly selected target patches. Table \ref{tab:b1b3} lists the tissue classification errors of the three classifiers and the proxy ${\cal A}$-distance between the source and target patches before ({\sc raw}) and after ({\sc rep}) applying {\sc mrai-net}. The whole experiment was repeated 10 times and the average error rate is reported with the standard error of the mean between brackets.

\begin{table}[h!]
\centering
\caption{Manually versus randomly selecting 1 target patch per tissue from 1 subject. (Left) Tissue classification error is reported for {\sc mrai-net} (linear classifier after {\sc mrai-net}), {\sc source} (supervised CNN trained on source patches and 1 target patch per tissue), and {\sc target} (supervised CNN trained on 1 target patch per tissue) tested on the target test data. (Right) Proxy ${\cal A}$-distance between the original source and target patches ({\sc raw}) and the source and target patches after applying {\sc mrai-net} ({\sc rep}).}
\label{tab:b1b3}
\begin{tabular}{ l | c c c}
		& {\sc source} & {\sc mrai-net} & {\sc target} \\
\hline 
 manual & 0.631 (.02) & 0.223 (.01) & 0.613 (.01) \\ 
 random & 0.667 (.02) & 0.250 (.02) & 0.610 (.06)
\end{tabular}
\hfill
\begin{tabular}{ | c c }
{\sc raw} & {\sc rep} \\
\hline 
 1.88 (.01) & 0.26 (.05)  \\ 
 1.91 (.01) & 0.41 (.06)  
\end{tabular}
\end{table}

\begin{figure}[h!]
\centering
\includegraphics[width=.28\textwidth]{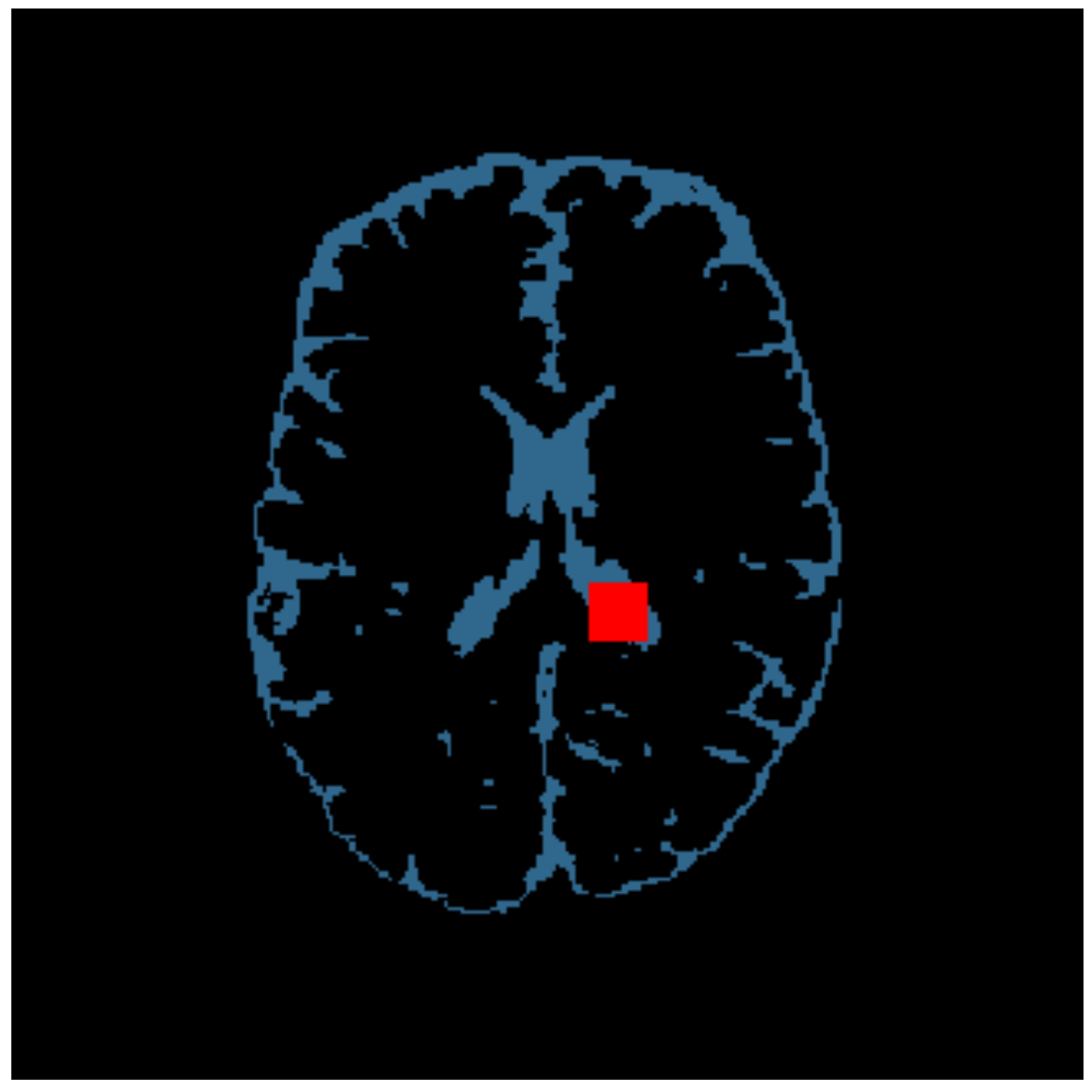} \hfill 
\includegraphics[width=.28\textwidth]{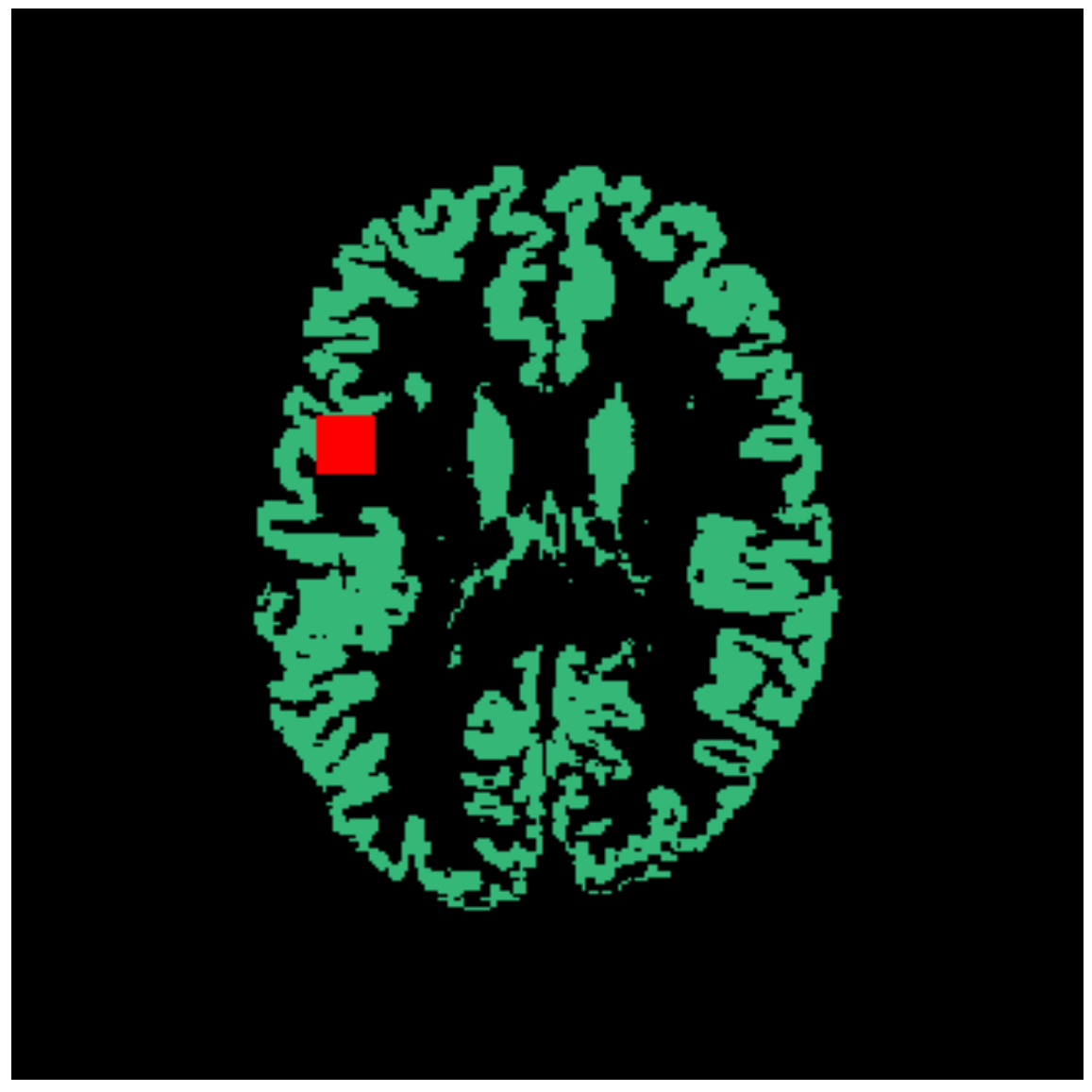} \hfill 
\includegraphics[width=.28\textwidth]{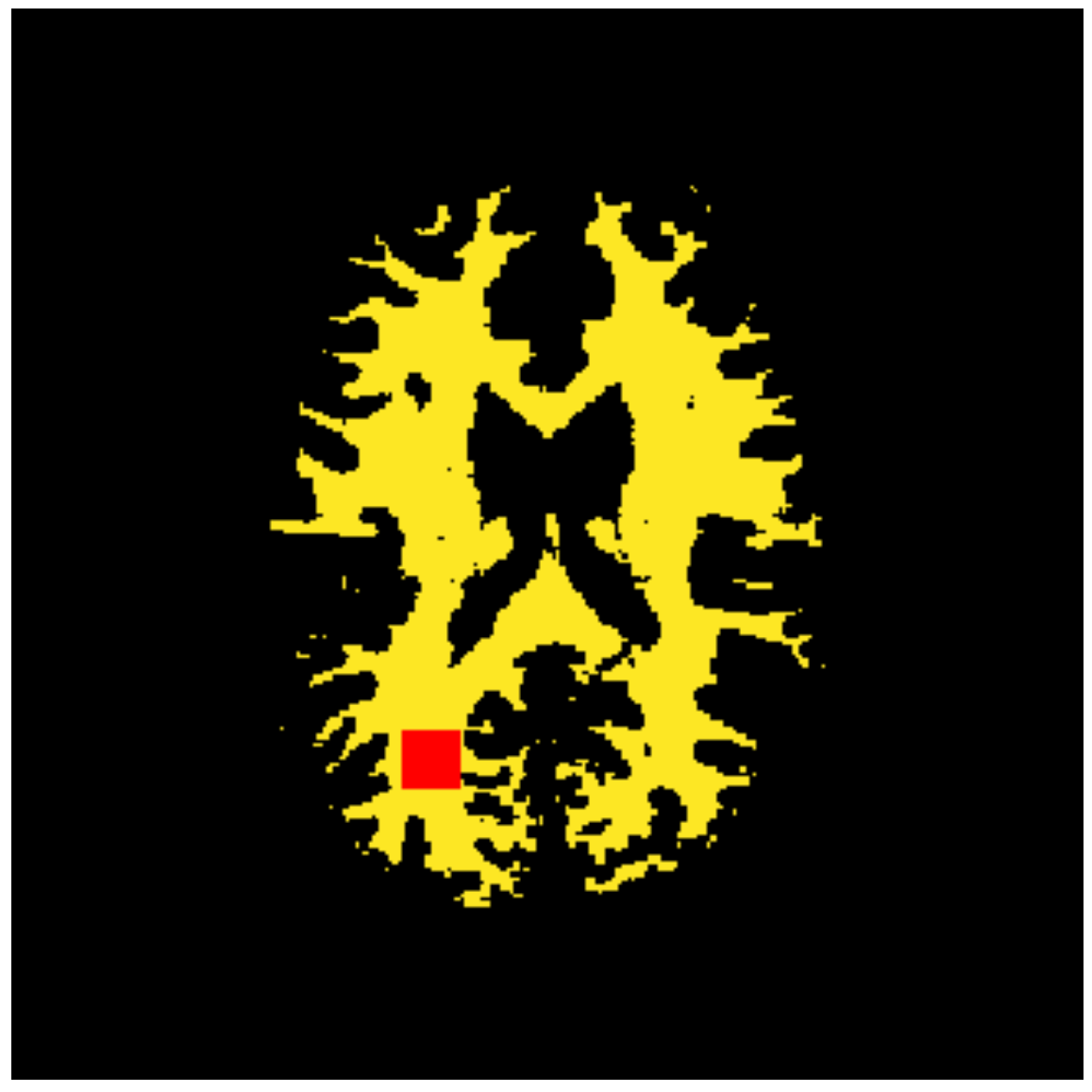} 
\caption{Locations of the manually selected target patches (red squares): Blue = cerebrospinal fluid, green = gray matter, yellow = white matter.}
\label{fig:manual}
\end{figure}

Figure \ref{fig:manual} displays the manually selected patches and their position within the image. For both the {\sc source}  and {\sc target} classifier, one target patch per tissue is insufficient to achieve good tissue classification performance (\ref{tab:b1b3} (top row): 0.631 and 0.613). However, the {\sc mrai-net} classifier shows considerably better performance (0.223), using only one target patch per tissue. The proxy ${\cal A}$-distance also drops from near perfect separability (1.88) to near invariance (0.26). Randomly selecting (10 repeats) 1 target patch per tissue (Table \ref{tab:b1b3} (bottom row)), shows worse performance of the {\sc mrai-net} classifier, for both the classification error (0.250) as well as the ${\cal A}$-distance (0.41). Suggesting that purposive (information rich) sampling beats random sampling in this case.

\subsection{Experiment 2: Multiple training target samples per tissue} \label{sec:MultSamples}
The second experiment tests the performance when adding more target training samples, for both simulated (Brainweb3.0T) and real patient data (MRBrainS). We set-up the following sub-experiments:
\begin{itemize}
\item [] 2.1) Experiment on simulated data with two different acquisition protocols (Source: Brainweb1.5T, Target: Brainweb3.0T).
\item [] 2.2) Experiment on 1.5T simulated data and 3.0T real data (Source: Brainweb1.5T, Target: MRBrainS). 
\item [] 2.3) Experiment on 3.0T simulated data and 3.0T real data (Source: Brainweb3.0T, Target: MRBrainS). 
\end{itemize}

\begin{figure}[h!]
\centering
\begin{small} {\bf Distance between source and target} \hfill {\bf Tissue classification target test data} \end{small} \\ \vspace{5px}
\includegraphics[width=.9\textwidth]{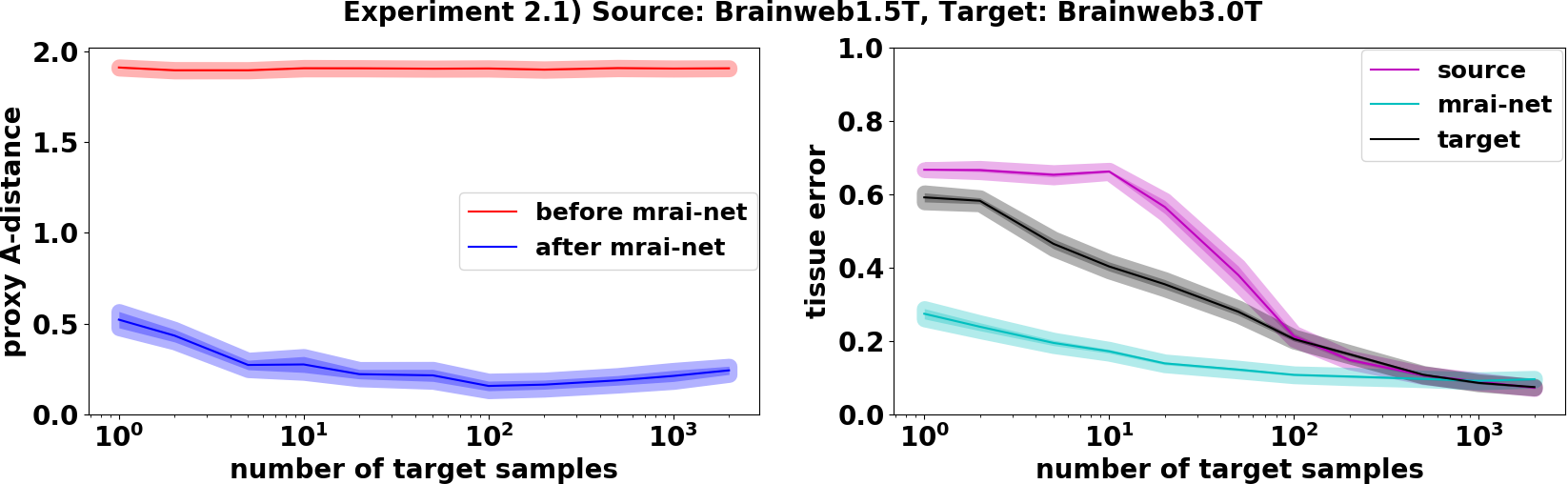} \\ \vspace{5px}
\includegraphics[width=.9\textwidth]{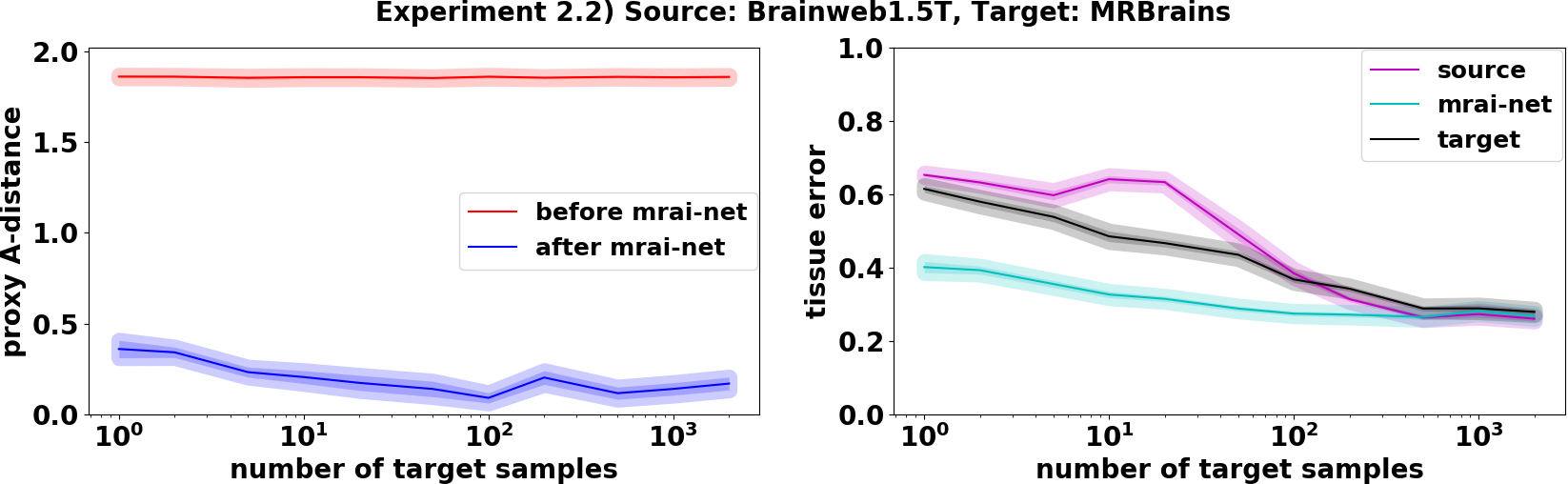} \\ \vspace{5px}
\includegraphics[width=.9\textwidth]{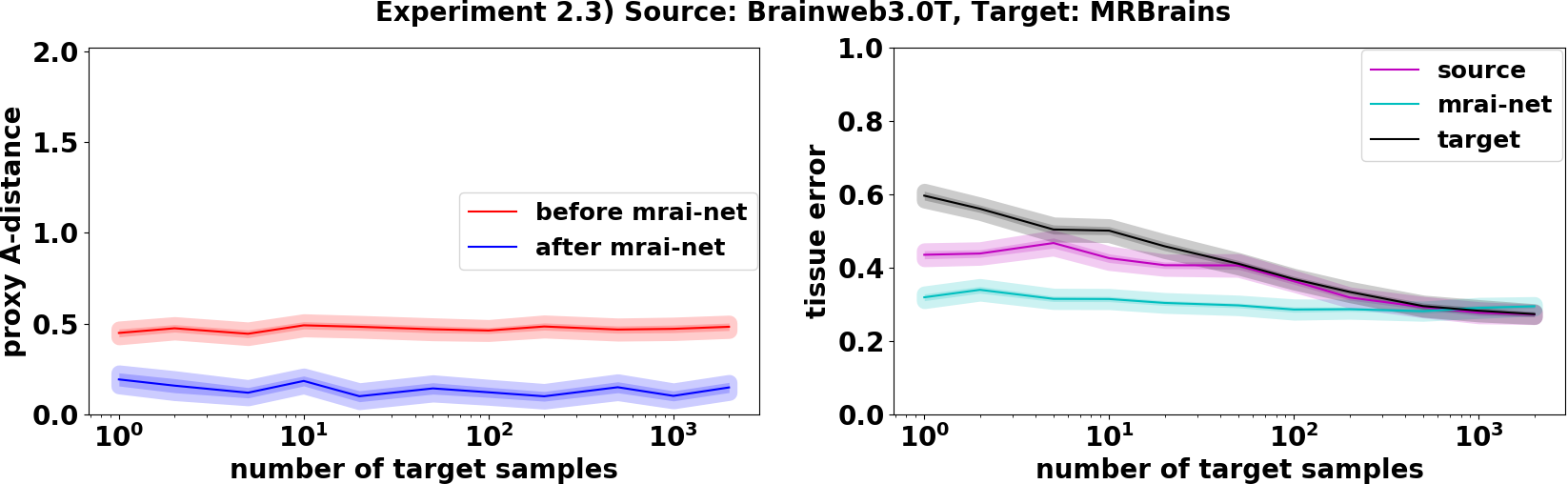} 
\caption{Graphs showing the effect of adding labeled samples from the target scanner for training the networks. (Left) Proxy A-distance between source and target scanner patches before (red) and after (blue) learning the {\sc mrai} representation (smaller distance is more acquisition-invariance). (Right) Tissue classification error for the three classifiers {\sc source} (supervised CNN trained on patches from source and target), {\sc {\sc mrai-net}} (supervised SVM trained on the source and target data mapped to {\sc mrai-net}'s representation) and {\sc target} (supervised CNN trained on target patches). Note that when the proxy ${\cal A}$-distance between the source and target data before {\sc mrai-net} is small (red line exp 2.3), the source data is representative of the target data (both 3T data), and the source tissue classifier (purple) shows better performance than using the target tissue classifier (cyan) with a small amount of target samples. However, if the proxy ${\cal A}$-distance is large (exp 2.1 and 2.2) before {\sc mrai-net} (red line), the source tissue classifier (purple) shows worse performance than the target tissue classifier (cyan) with a small amount of target samples, since the source data (1.5T) is not representative of the target data (3T).}
\label{fig:results_b1b3}
\end{figure}

Each of these experiments is repeated 50 times. Figure \ref{fig:results_b1b3} shows the performance (both tissue classification error as well as proxy ${\cal A}$-distance) as a function of the number of used target training samples. The average error (solid line) and the standard error of the mean (line thickness) is shown, ranging from using 1 target patch up to more than 1000 target patches per tissue for training both the supervised CNNs ({\sc source}, {\sc target}) as well as the {\sc mrai-net} followed by a linear classifier ({\sc mrai-net}).

Figure \ref{fig:results_b1b3} (left) shows the proxy ${\cal A}$-distance between the source and target samples for all three experiments. The proxy ${\cal A}$-distance for experiments 2.1 and 2.2 shows that in the original representation (raw; red line), the source and target distributions lie far apart (proxy ${\cal A}$-distance approaches $2$). This illustrates the difference in acquisition protocol (1.5T versus 3.0T). After applying {\sc mrai-net} (rep; blue line) the proxy ${\cal A}$-distance drops drastically (approaches $0$) showing that the network managed to learn an MR-acquisition invariant representation. Adding more target training samples improves the invariance up to about 100 samples, but the proxy ${\cal A}$-distance is already quite low after only using 1 target sample per tissue type for training. In experiment 2.3 the proxy ${\cal A}$-distance before applying {\sc mrai-net} ({\sc raw}) is already much lower than in the previous two experiments (around $0.5$), this illustrates that the acquisition protocols are more similar to begin with (both 3.0T). The main difference between the distributions presumably results from simulated versus real data, since not all factors of acquisition variation are included in the simulations, most notably partial volume (0.96x0.96x3mm voxels in MRBrainS versus no partial volume in Brainweb). However, after applying {\sc mrai-net} the proxy ${\cal A}$-distance is reduced further (approaches $0$), again showing that {\sc mrai-net} is able to learn an MR-acquisition invariant representation ({\sc rep}) on this data, even for simulated and real data. Note that the MRBrainS data adds other modes of variation in terms of pathology and age in comparison to the Brainweb healthy adults, which could influence the tissue classification performance. 

\begin{figure}[h!]
\centering
\begin{subfigure}[t]{.24\textwidth}
	\includegraphics[width=.98\textwidth]{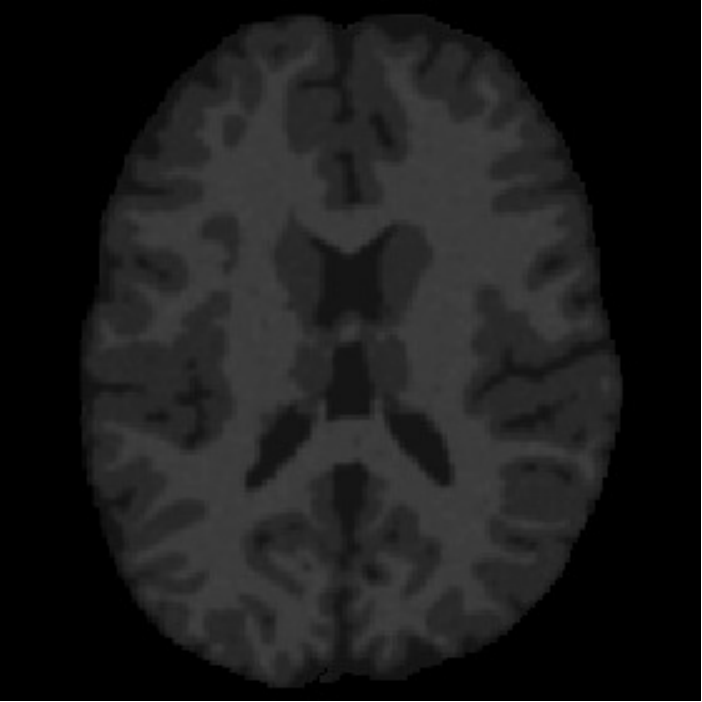}
    \caption{Scan}
\end{subfigure}
\begin{subfigure}[t]{.24\textwidth}
	\includegraphics[width=.98\textwidth]{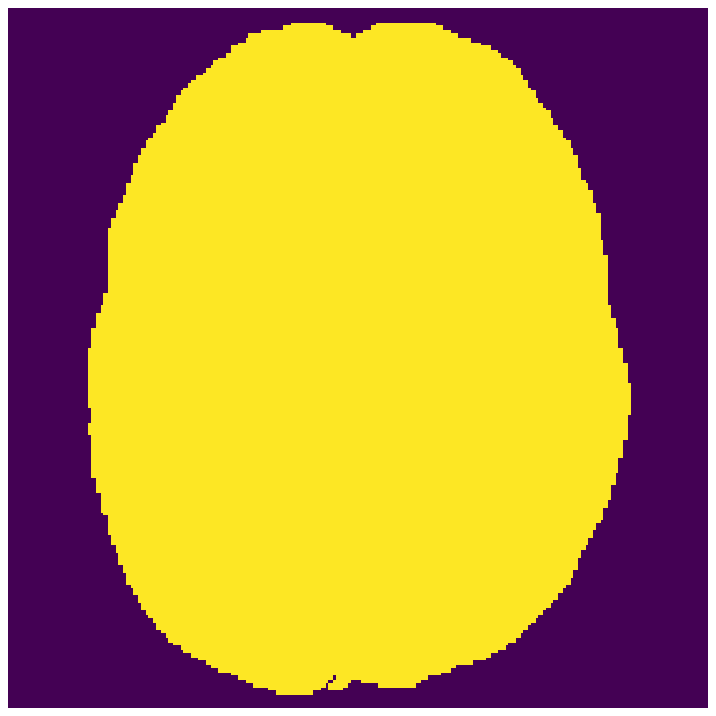}
    \caption{{\sc source} (1 TP)}
\end{subfigure}
\begin{subfigure}[t]{.24\textwidth}
	\includegraphics[width=.98\textwidth]{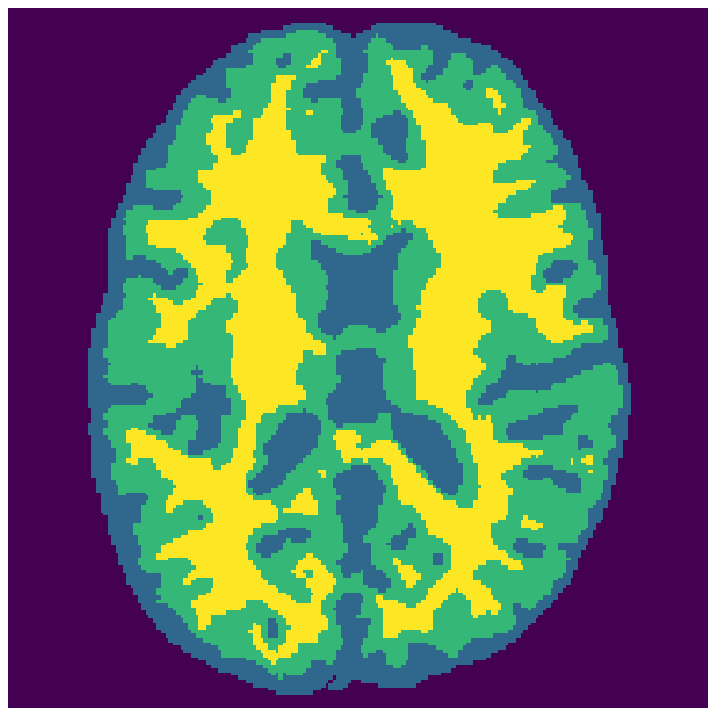}
    \caption{{\sc mrai-net} (1 TP)}
\end{subfigure} 
\begin{subfigure}[t]{.24\textwidth}
	\includegraphics[width=.98\textwidth]{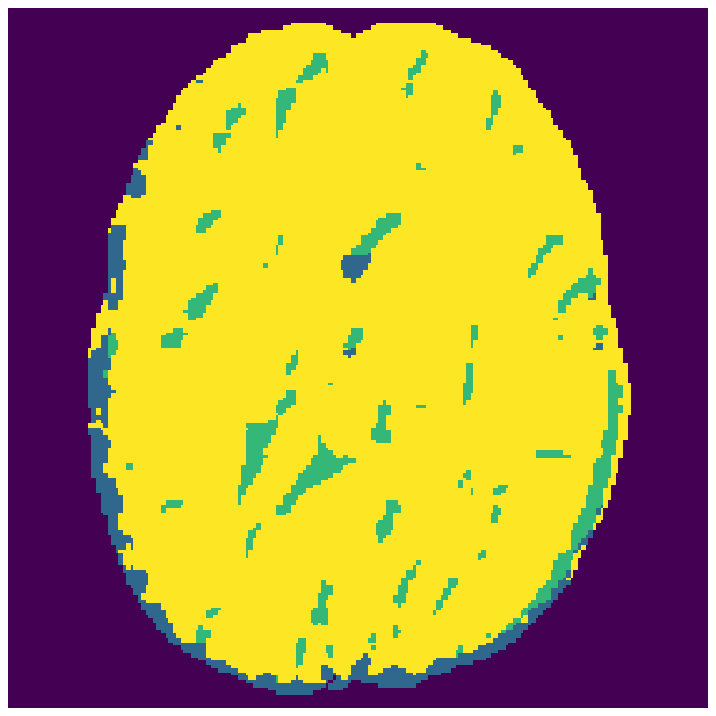}
    \caption{{\sc target} (1 TP)}
\end{subfigure} \\

\begin{subfigure}[t]{.24\textwidth}
	\includegraphics[width=.98\textwidth]{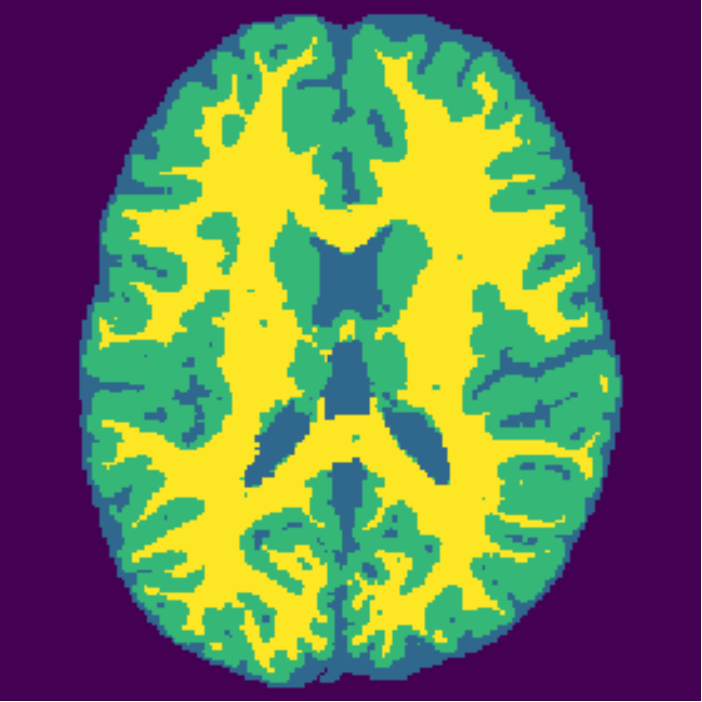}
    \caption{Ground truth}
\end{subfigure}
\begin{subfigure}[t]{.24\textwidth}
	\includegraphics[width=.98\textwidth]{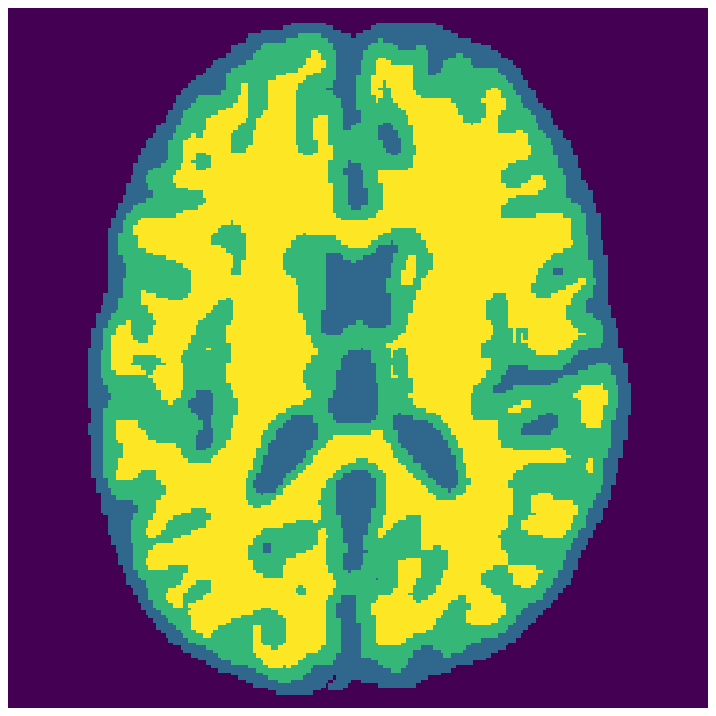}
    \caption{{\sc source} (100 TPs)}
\end{subfigure}
\begin{subfigure}[t]{=.24\textwidth}
	\includegraphics[width=.98\textwidth]{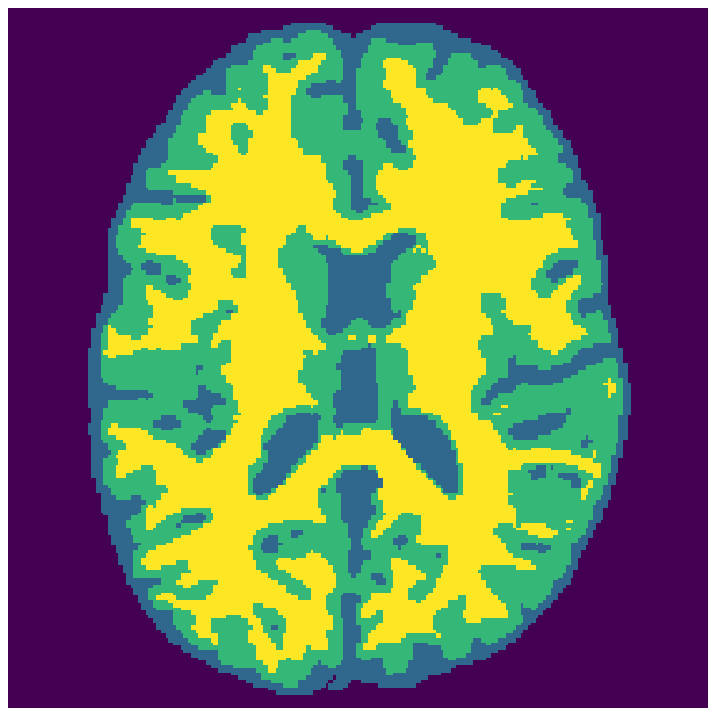}
    \caption{{\sc mrai-net} (100 TPs)}
\end{subfigure}
\begin{subfigure}[t]{.24\textwidth}
	\includegraphics[width=.98\textwidth]{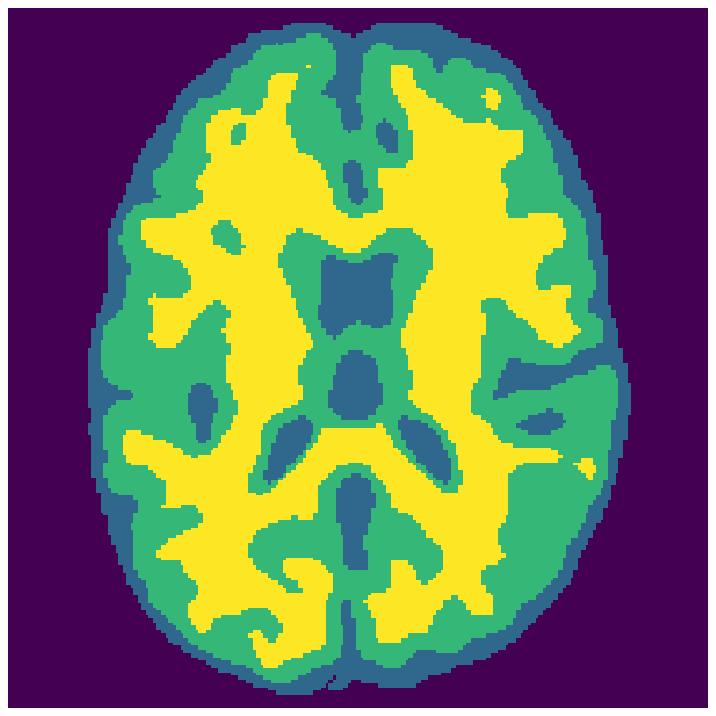}
    \caption{{\sc target} (100 TPs)}
\end{subfigure}
\caption{Example brain tissue segmentations into white matter (yellow), gray matter (green) and cerebrospinal fluid (blue) for experiment 2.1 (Source: Brainweb1.5T, Target: Brainweb3.0T). A simulated MRI scan of a test subject from Brainweb3.0T (a) is shown, with corresponding ground truth segmentation (e), and the results of applying the {\sc source} (b,f), {\sc target} (d,h) and proposed {\sc mrai-net} (c,g) classifiers, with either 1 or 100 target patches per tissue type used for training the classifiers (Figure \ref{fig:results_b1b3}).}
\label{fig:preds_b1b3}
\end{figure}

Figure \ref{fig:results_b1b3} (right) shows the tissue classification error for all three experiments. If the proxy ${\cal A}$-distance between the source and target distribution is high (experiment 2.1 and 2.2), and when using only one target sample per tissue, the {\sc source} classifier that uses both the source data and target data for training shows worse performance than the one that uses only the target data ({\sc target}); an error of 0.667 versus 0.591, respectively. Even when adding more target samples for training, the results show that it is more beneficial to train a supervised classifier on the target data alone, instead of on both the source and target data; using 10 target samples for training, {\sc source} achieves an error of 0.662 versus an error of 0.403 for {\sc target}. The {\sc source} classifier is focused on its source samples, which in this case are not informative of the target data. Given enough target samples, however, {\sc source} starts to shift focus towards its target data and starts to match the performance of {\sc target}: for 100 target samples, errors of 0.213 versus 0.205 respectively. If the proxy ${\cal A}$-distance between the source and target distributions is low (distributions are more similar; experiment 2.3), using the source data for training is beneficial; for 1 target sample per tissue {\sc source} achieves an error of 0.435 and {\sc target} an error of 0.596. In this case, the source samples are more representative of the target data and are aiding the classifier. In general, the {\sc mrai-net} classifier outperforms both the {\sc source} and {\sc target} classifiers: an error of 0.269 for 1 sample, 0.175 for 10 samples and 0.111 for 100 samples. {\sc mrai-net}'s representation ensures that the source and target samples are more similar and that the source samples can be effectively used for training.

\begin{figure}[h!]
\centering
\begin{subfigure}[t]{.24\textwidth}
	\includegraphics[width=.98\textwidth]{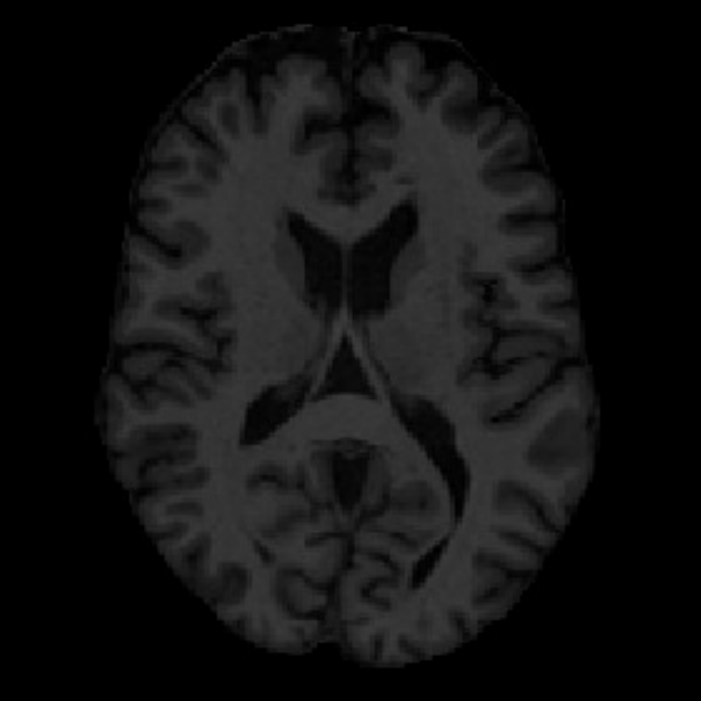}
    \caption{Scan}
\end{subfigure}
\begin{subfigure}[t]{.24\textwidth}
	\includegraphics[width=.98\textwidth]{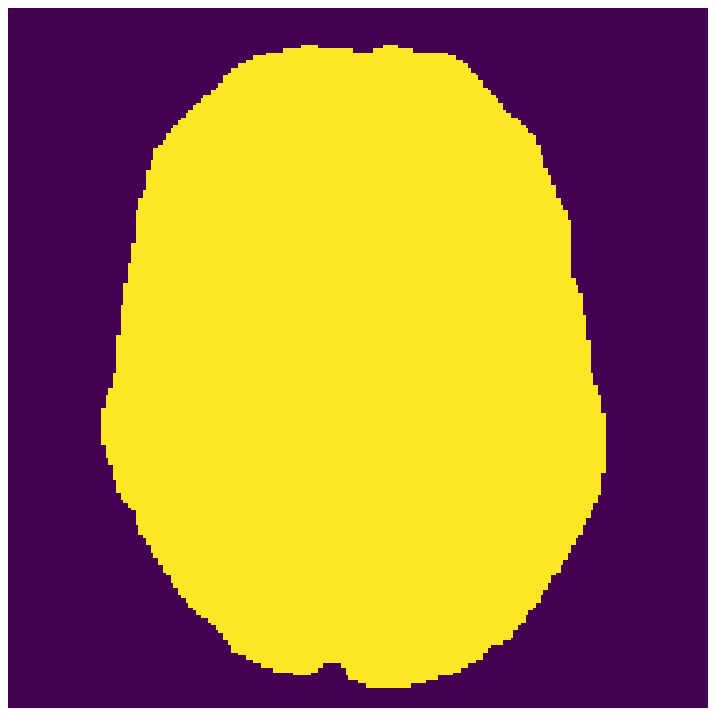}
    \caption{{\sc source} (1 TP)}
\end{subfigure}
\begin{subfigure}[t]{.24\textwidth}
	\includegraphics[width=.98\textwidth]{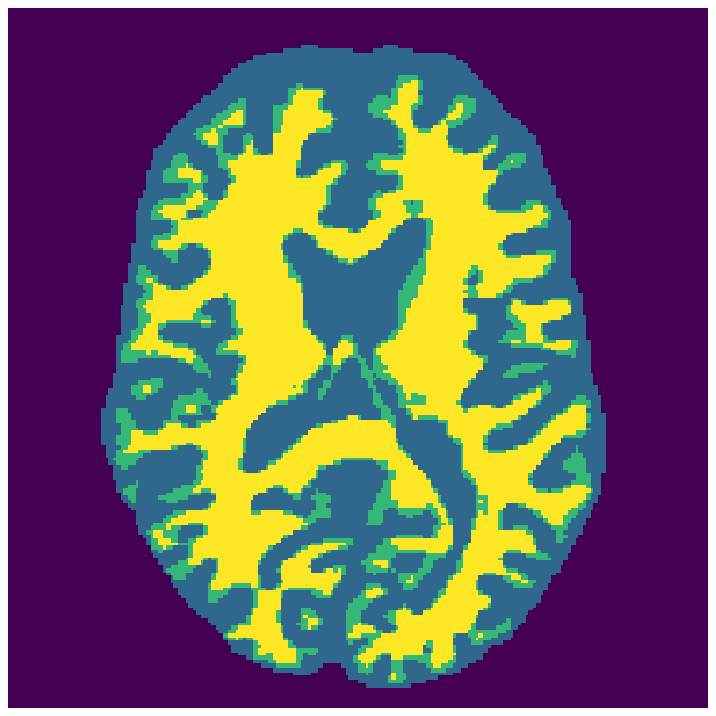}
    \caption{{\sc mrai-net} (1 TP)}
\end{subfigure} 
\begin{subfigure}[t]{.24\textwidth}
	\includegraphics[width=.98\textwidth]{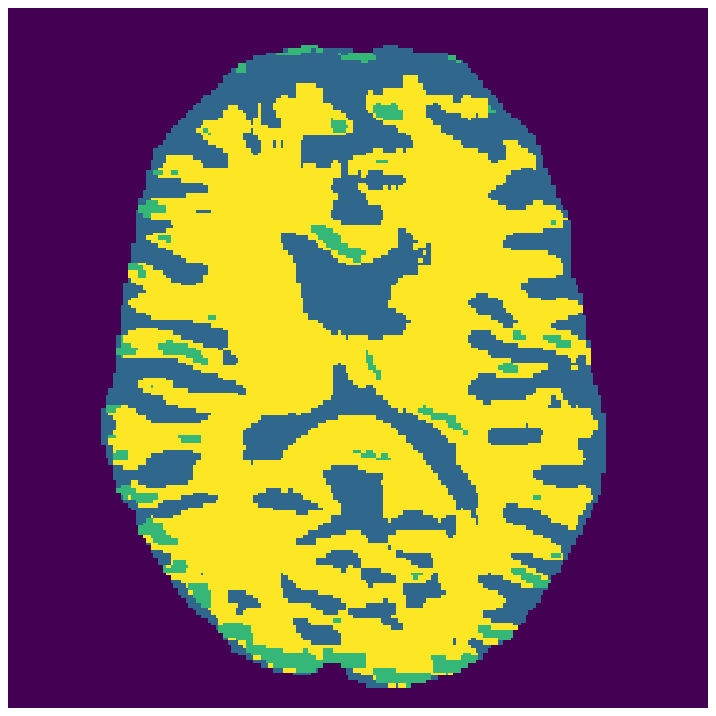}
    \caption{{\sc target} (1 TP)}
\end{subfigure} \\

\begin{subfigure}[t]{.24\textwidth}
	\includegraphics[width=.98\textwidth]{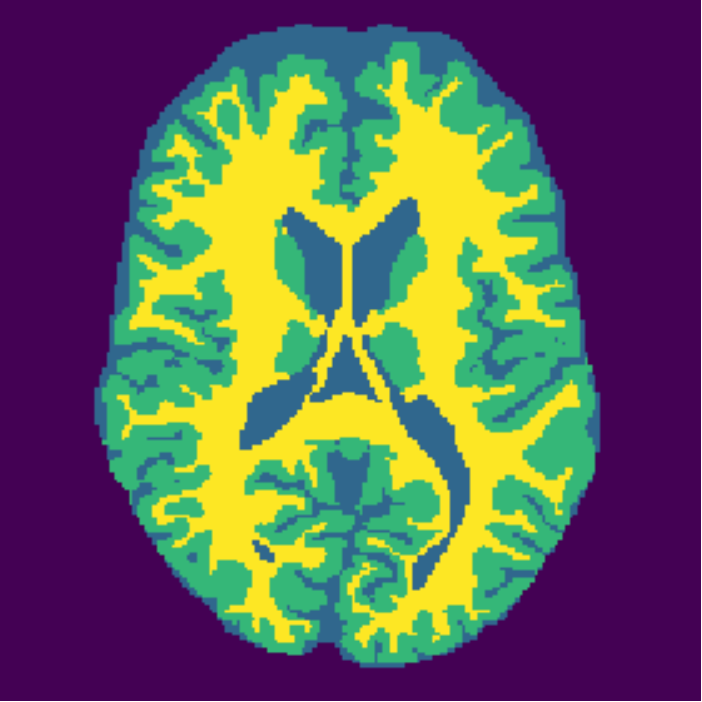}
    \caption{Ground truth}
\end{subfigure}
\begin{subfigure}[t]{.24\textwidth}
	\includegraphics[width=.98\textwidth]{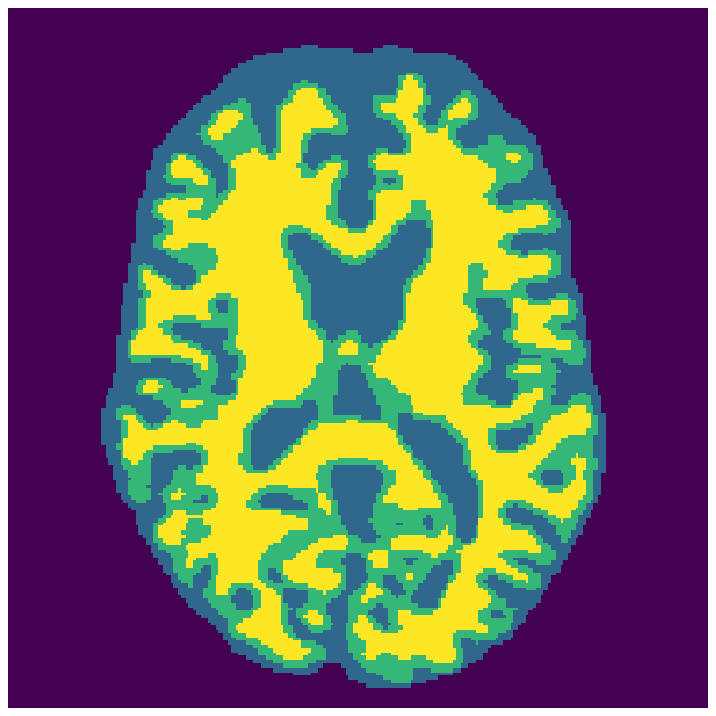}
    \caption{{\sc source} (100 TPs)}
\end{subfigure}
\begin{subfigure}[t]{=.24\textwidth}
	\includegraphics[width=.98\textwidth]{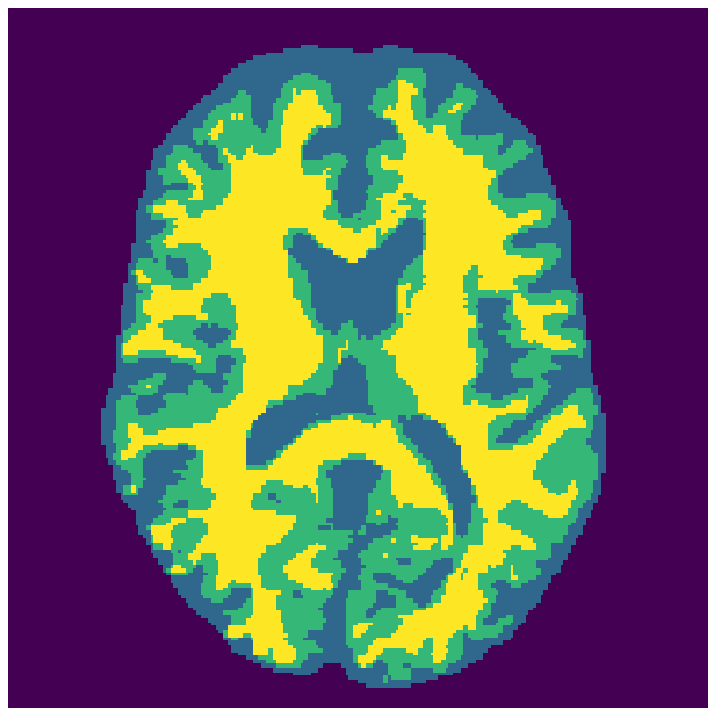}
    \caption{{\sc mrai-net} (100 TPs)}
\end{subfigure}
\begin{subfigure}[t]{.24\textwidth}
	\includegraphics[width=.98\textwidth]{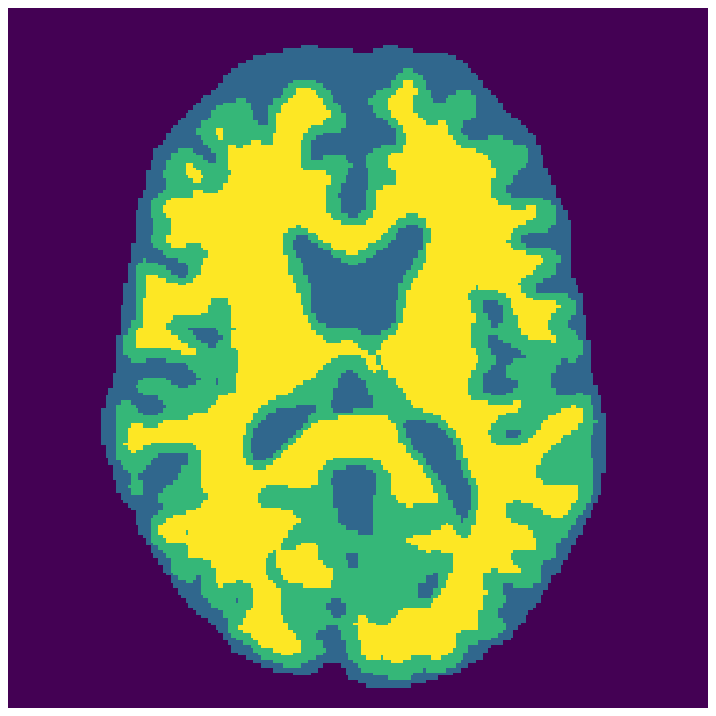}
    \caption{{\sc target} (100 TPs)}
\end{subfigure}
\caption{Example brain tissue segmentations into white matter (yellow), gray matter (green) and cerebrospinal fluid (blue) for experiment 2.2 (Source: Brainweb1.5T, Target: MRBrainS). A simulated MRI scan of a test subject from MRBrainS (a) is shown, with corresponding ground truth segmentation (e), and the results of applying the {\sc source} (b,f), {\sc target} (d,h) and proposed {\sc mrai-net} (c,g) classifiers, with either 1 or 100 target patches per tissue type used for training the classifiers (Figure \ref{fig:results_b1b3}).}
\label{fig:preds_b1mb}
\end{figure}

\begin{figure}[h!]
\centering
\begin{subfigure}[t]{.24\textwidth}
	\includegraphics[width=.98\textwidth]{mrbrains_I04.eps}
    \caption{Scan}
\end{subfigure}
\begin{subfigure}[t]{.24\textwidth}
	\includegraphics[width=.98\textwidth]{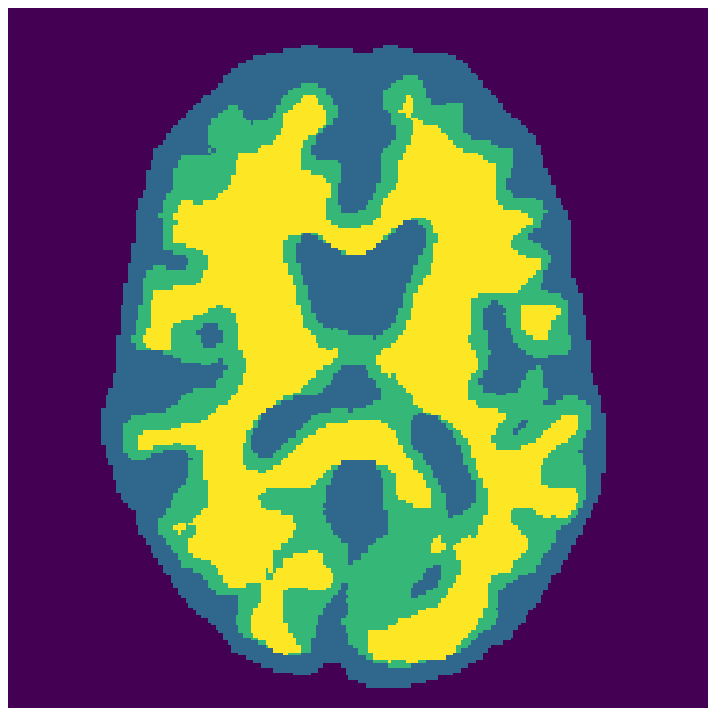}
    \caption{{\sc source} (1 TP)}
\end{subfigure}
\begin{subfigure}[t]{.24\textwidth}
	\includegraphics[width=.98\textwidth]{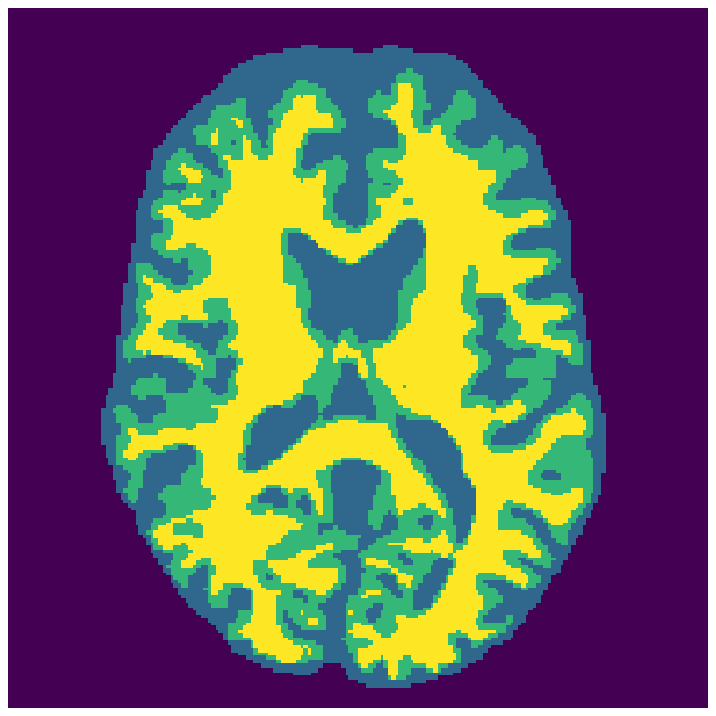}
    \caption{{\sc mrai-net} (1 TP)}
\end{subfigure} 
\begin{subfigure}[t]{.24\textwidth}
	\includegraphics[width=.98\textwidth]{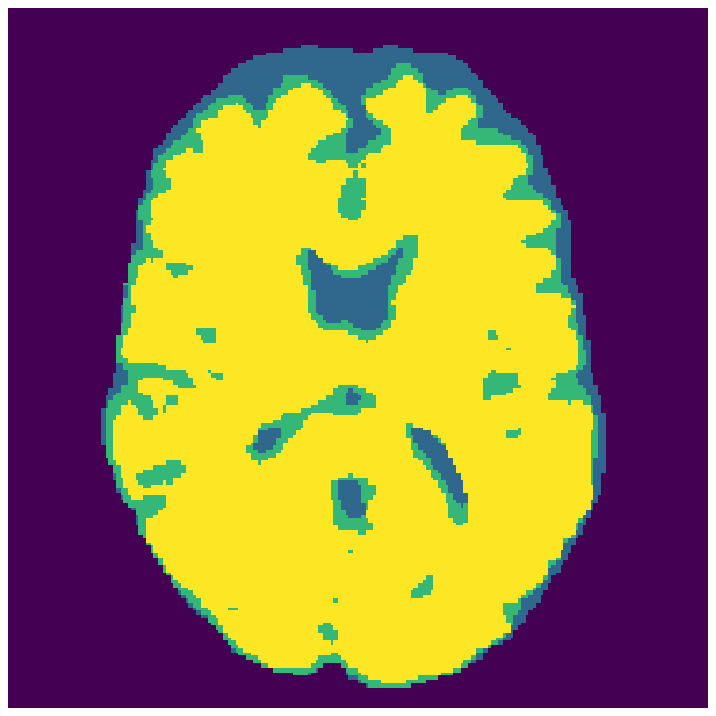}
    \caption{{\sc target} (1 TP)}
\end{subfigure} \\

\begin{subfigure}[t]{.24\textwidth}
	\includegraphics[width=.98\textwidth]{mrbrains_L04.eps}
    \caption{Ground truth}
\end{subfigure}
\begin{subfigure}[t]{.24\textwidth}
	\includegraphics[width=.98\textwidth]{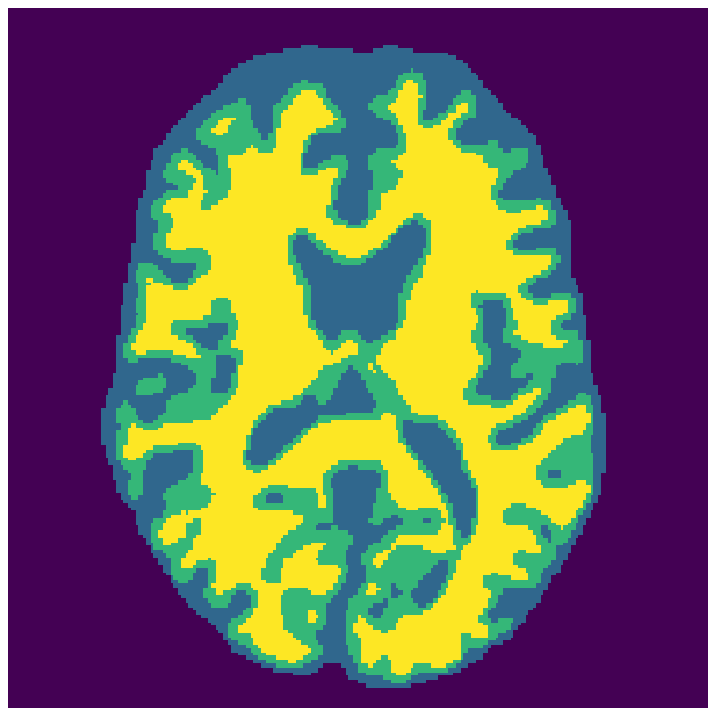}
    \caption{{\sc source} (100 TPs)}
\end{subfigure}
\begin{subfigure}[t]{=.24\textwidth}
	\includegraphics[width=.98\textwidth]{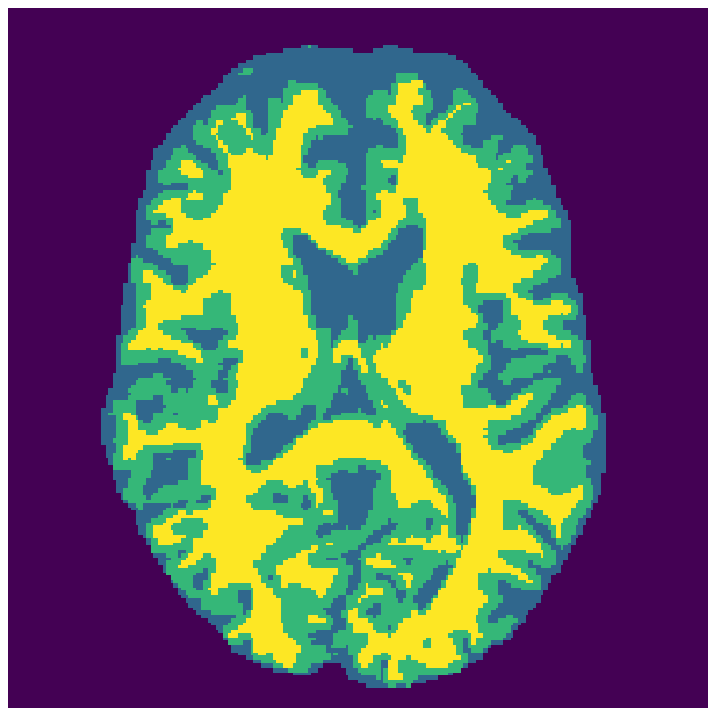}
    \caption{{\sc mrai-net} (100 TPs)}
\end{subfigure}
\begin{subfigure}[t]{.24\textwidth}
	\includegraphics[width=.98\textwidth]{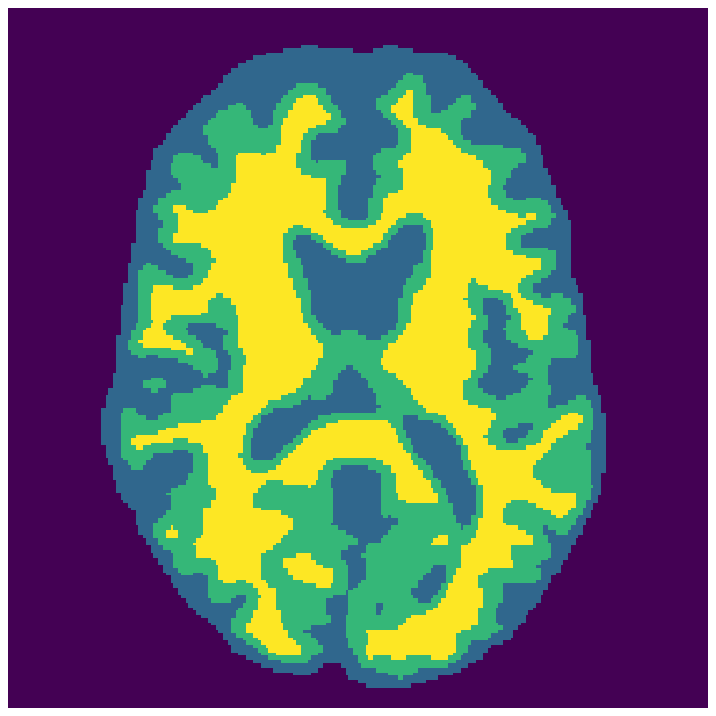}
    \caption{{\sc target} (100 TPs)}
\end{subfigure}
\caption{Example brain tissue segmentations into white matter (yellow), gray matter (green) and cerebrospinal fluid (blue) for experiment 2.3 (Source: Brainweb3.0T, Target: MRBrainS). A simulated MRI scan of a test subject from MRBrainS (a) is shown, with corresponding ground truth segmentation (e), and the results of applying the {\sc source} (b,f), {\sc target} (d,h) and proposed {\sc mrai-net} (c,g) classifiers, with either 1 or 100 target patches per tissue type used for training the classifiers (Figure \ref{fig:results_b1b3}.}
\label{fig:preds_b3mb}
\end{figure}

Examples of the segmentation results on one of the target test images are shown in Figure \ref{fig:preds_b1b3} for experiment 2.1, Figure \ref{fig:preds_b1mb} for experiment 2.2, and Figure \ref{fig:preds_b3mb} for experiment 2.3. Examples are shown after using 1 target patch per tissue for training, and after using 100 target patches per tissue for training. The results show that only the {\sc mrai-net} classifier is able to predict a segmentation that approaches the ground truth with only 1 target patch per tissue for training (error for experiment 2.1 = 0.269, experiment 2.2 = 0.403, experiment 2.3 = 0.320), while the {\sc source} and {\sc target} classifiers cannot ({\sc source} error for experiment 2.1 = 0.667, experiment 2.2 = 0.653, experiment 2.3 = 0.435; {\sc target} error for experiment 2.1: 0.591, experiment 2.2: 0.614, experiment 2.3 = 0.596). After using 100 patches the {\sc source} and {\sc target} classifiers can predict a gross segmentation of WM, GM and CSF ({\sc source} error for experiment 2.1 = 0.213, experiment 2.2 = 0.384, experiment 2.3 = 0.363; {\sc target} error for experiment 2.1: 0.205, experiment 2.2: 0.368, experiment 2.3 = 0.368), but the {\sc mrai-net} classifier prediction shows more details and a lower tissue classification error (error for experiment 2.1 = 0.111, experiment 2.2 = 0.276, experiment 2.3 = 0.284).

\subsection{Experiment 3: number of network parameters} \label{sec:exp-numparams}
Setting neural network hyperparameters, such as the number of convolution kernels to use, is always a tricky issue. The optimal parameter is different for each dataset, which means there are no easy defaults. In order to get some insight into the behavior of the network for different choices of hyperparameters, we performed an additional experiment. We used experiment 2.1's setting: Brainweb1.5T as source and Brainweb3.0T as target.

\begin{figure}[h!]
\centering 
\includegraphics[width=0.95\textwidth]{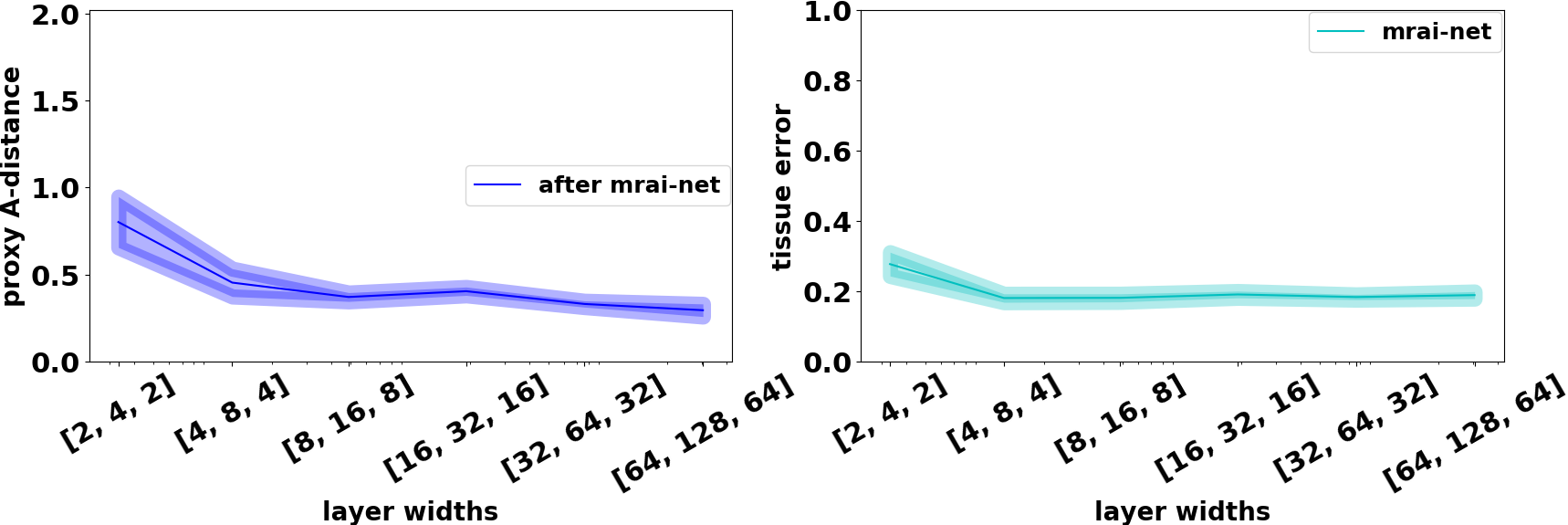}
\caption{{\sc mrai-net}'s performance as a function of layer widths. (Left) The proxy ${\cal A}$-distance. (Right) The tissue classification error obtained through a linear classifier trained on data in {\sc mrai-net}'s representation. Both graphs show a slow gain in performance as the number of parameters grows.}
\label{fig:numparams}
\end{figure}

{\sc mrai-net} has three layers with parameters: a convolution layer and two dense layers. We varied the number of kernels in the convolution layer and the number of nodes in the dense layers. We use the following sets of hyperparameters: [2 kernels, 4 nodes, 4 nodes], [4 kernels, 8 nodes, 4 nodes], [8 kernels, 16 nodes, 8 nodes], [16 kernels, 32 nodes, 16 nodes], [32 kernels, 64 nodes, 32 nodes] and [64 kernels, 128 nodes, 64 nodes] (i.e. the layer widths double each time). The total number of parameters are 322, 1254, 4874, 19218, 76322, and 304194, respectively. We used 10 labeled target patches per classes, from which we generated 18000 pairs of patches. The network was trained for 320 epochs and the experiment was repeated 20 times to obtain standard errors of the means. Figure \ref{fig:numparams} shows the results: the left figure looks at the proxy ${\cal A}$-distance as a function of the number of parameters and the right figure looks at the tissue classification error of a linear classifier trained on the resulting representation. For the proxy ${\cal A}$-distance, the graphs show a steady decrease in distance and then roughly levels off after [8, 16, 8]. This result indicates that an extremely wide {\sc mrai-net} (i.e. [64, 128, 64]) will still be able to reduce acquisition variation. As for the tissue classification error, the thin network (i.e. [2, 4, 2]) starts out with a average error rate of 0.28 (underfitting) and drops immediately to 0.18 for [4, 8, 4]. Afterwards, it slowly increases to 0.19. This indicates that the network is not overfitting too drastically yet, which is probably due to the regularization (see Section \ref{sec:reg}). However, the graph does indicate that its error rate will go up if the number of parameters is increased further. 

\subsection{Experiment 4: effect of the margin parameter} \label{exp:margin}
\begin{figure}[h!]
\begin{subfigure}[b]{.6\textwidth}
	\centering
	\includegraphics[width=.95\textwidth]{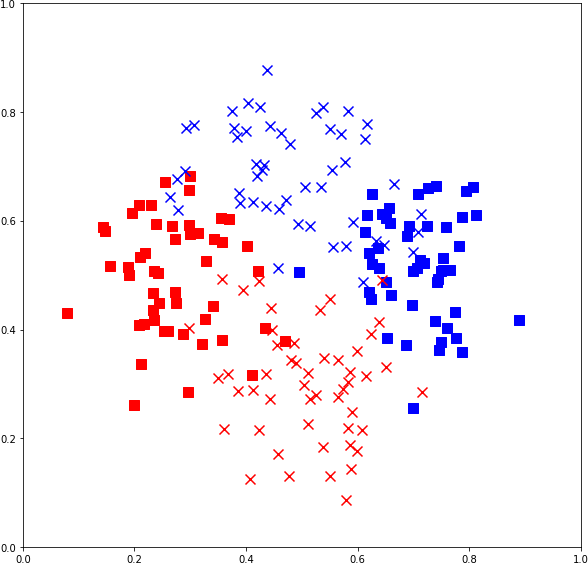} 
\end{subfigure}
\begin{subfigure}[b]{.38\textwidth}
	\centering
	\includegraphics[width=.97\textwidth]{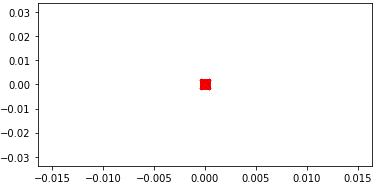} \\
	\includegraphics[width=.97\textwidth]{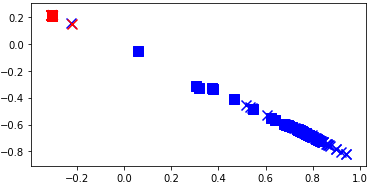} \\
	\includegraphics[width=.97\textwidth]{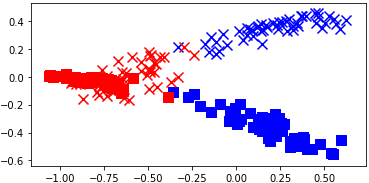} 
\end{subfigure}
\caption{Effect of the margin hyperparameter. (Left) Two synthetic binary data sets, with markers indicating scanners and colors tissues. (Right) Representation found by a network with a margin of 0 (top), a margin of 1 (middle) and a margin of 10 (bottom).}
\label{fig:margins}
\end{figure}

The margin parameter $m$ in the dissimilar loss function, $\ell_{\text{dis}}( f \given a,b) = \max(0,m-\|f(a) - f(b) \|_p)$, is important as it balances the actions of pushing and pulling between pairs. For small values, $\ell_{\text{dis}}$ will be much smaller than $\ell_{\text{sim}}$ and the network will focus on pulling pairs together. For large values, $\ell_{\text{dis}}$ will always be much larger than $\ell_{\text{sim}}$ and network will focus on pushing pairs apart. Figure \ref{fig:margins} plots a synthetic data setting with the outcome of using three different values for the margin parameter. The left figure shows two synthetic 2-dimensional data sets, one with red versus blue crosses and the other with red versus blue squares. The right figures show validation samples fed through three networks with different values for the margin parameter. Firstly, the right top figure displays the result of using a margin parameter of $0$: the network does not suffer \emph{any} loss by making pairs of samples of different tissues too similar and consequently maps everything to a single point. Secondly, the right middle figure shows an appropriate choice for the margin, where the two data sets overlap and where red and blue points are separated. Lastly, the right bottom figure shows what happens when a large margin parameter is used: it focuses almost entirely on separating red versus blue and is not making the data sets more similar. 

\paragraph{} Additionally, the optimal value for the margin parameter is affected by the number of similar versus dissimilar pairs. If there are twice as many similar pairs, then their loss will be twice as large as well and the network will focus more on pulling pairs together. Overall, the more similar pairs there are, the larger the margin parameter will need to be.

\section{Discussion}
In this paper, we proposed a method to learn an MR acquisition invariant ({\sc mrai}) representation that preserves the variation between brain tissues for segmentation. Once the representation is learned using {\sc mrai-net}, any supervised classification model that uses feature vectors can be used to classify the brain tissues. The proposed method addresses the problem that the difference between scans acquired with two different MRI scanners or protocols can be so large that scans from one scanner are not representative of scans from another scanner. This difference does not affect assessment by human vision (e.g. radiologists can perform diagnostic work-up on both), but it does affect computer vision. To get insight into the difference between scans and to assess the performance of {\sc mrai-net} to reduce this difference (achieve invariance), the proxy ${\cal A}$-distance measure between source and target patches was used. The experiments (Figure \ref{fig:results_b1b3}) show that this is a good measure to determine the difference between source and target acquisition, and might be used to predict classifier performance of a source classifier. Note that this measure does not require any tissue labels, and can thus be used as a general measure of distance between scanners. It merely requires source patches to be labeled as source, and target patches to be labeled as target. When the proxy ${\cal A}$-distance is low (Figure \ref{fig:results_b1b3} bottom row) the source ({\sc source}) classifier outperforms the target ({\sc target}) classifier when a small number of target training patches are used. When the proxy ${\cal A}$-distance is large the {\sc target} classifier outperforms the {\sc source} classifier, even when one target training patch per tissue is used. This suggests that if the proxy ${\cal A}$-distance is large (source data is not representative of target data), a source classifier trained on the source data should not be applied to the target data. Ground truth labels on the source data that are labor-intensive to acquire can in this case not be used for the target data. However, since {\sc mrai-net} learns a representation that reduces the acquisition difference between source and target scanner the proxy ${\cal A}$-distance is drastically reduced. Therefore the {\sc mrai-net} classifier outperforms both the {\sc source} and {\sc target} classifiers, when a small number of target training samples is available, and leverages the source ground truth labels.\\
Due to the complexity of the problem addressed in this paper, simulated data was used to provide a proof of principle. Ideal real data would require the same subject to be scanned on different scanners with different protocols, after which the scans should be manually segmented to obtain the ground truth for both scans. However, inter-observer variability would add an extra layer of variation. To test {\sc mrai-net} on real data, the MRBrainS challenge data was used. Although, additional layers of variation include resolution, population and manual segmentation protocol, the experiments (Figure \ref{fig:results_b1b3}) show that the {\sc mrai-net} performance on real data follows the same pattern as its performance on simulated data, be it with a higher classification error due to additional factors of variation.\\ 
A limitation of the proposed method is that learning an {\sc mrai} representation with {\sc mrai-net}, will not necessarily work well on data sets with poor contrast between tissues. In that case, the network will both push and pull points in the overlap. Since these actions will mostly cancel each other out, the network will not be able to reduce acquisition-variation without sacrificing tissue variation, and vice versa. \\
Another limitation is that the proposed {\sc mrai-net} requires at least 1 sample per tissue from the target scanner. This is not an unreasonable request, as it is not hard to find at least 1 patch per tissue (Section \ref{sec:1sample}) in only one subject scanned with the target scanner. However, it may be possible to perform the similar/dissimilar labeling based on assumptions instead. For instance, if one assumes that the registration between two scans is accurate and that the subject-variation is not too large, then one could assume that target patches at certain locations are the same tissue as the source patches at these locations. Hence, those voxels could be used for the similarity-labeling process.\\
The proposed representation learning method could be used to reduce any type of variation, by adjusting the way that the similar and dissimilar pairs are defined. For example, registration, which can be viewed as variation in position, might be approached in a similar manner \cite{simonovsky2016deep}. Key is to identify the forms of variation, determine which variation should be preserved and which should be reduced, and to find a way to label them as similar or dissimilar accordingly.  

\section{Conclusion}
In this paper we addressed one of the major challenges of supervised voxel classification, i.e. generalization to data that is not representative of the training data. We provided a proof of principle for learning an MR acquisition invariant representation that reduces the variation between MRI scans acquired with different scanners or acquisition protocols, while preserving the variation between brain tissues. We showed that the proposed {\sc mrai-net} is able to learn an MR acquisition invariant representation (low proxy ${\cal A}$-distance), and outperform supervised convolution neural networks trained on patches from the source or target scanners for tissue classification, when little target training patches are available. By reducing the acquisition related variation using {\sc mrai-net}, the ground truth labels from the source data can be reused for the target data, since the source and target data are mapped to the same representation achieving generalization.

\section{Acknowledgements}
The research of A.M. Mendrik was financially supported by IMDI Grant 104002002 (Brainbox) from ZonMw, the Netherlands Organisation for Health Research and Development. W.M. Kouw was supported by the Netherlands Organization for Scientific Research (NWO; grant 612.001.301).

\begin{appendices}
\section{Nuclear Magnetic Resonance relaxation times} \label{app:NMR}
SIMRI requires NMR relaxation times for tissues based on particular magnetic static field strengths \cite{benoit2005simri}. We performed a literature study for the T1 and T2 relaxation times, the results of which are listed in Table \ref{sim_prop}. The proton density values $\rho$ stem from \cite{yoder2002mri}. The 3.0T CSF parameters were interpolated using an exponential function fit (\cite{kroeker1986analysis} justifies an exponential function based on physical properties). We equate connective tissue to glial matter (90\% of the brain's connective tissue system is glial matter\footnote{\url{http://www.neuroplastix.com/styled-2/page139/styled-42/brainsconnectivetissue.html}}).
\begin{table}[H]                                                               
\caption{NMR relaxation times for brain tissue (IT'IS database). \newline 
\textsuperscript{a} Glial matter values are unknown and are imputed with gray matter values.\newline 
\textsuperscript{b} T2 values for cortical bone are actually T2* values (UTE seq).\newline
\textsuperscript{c} Equated to glial matter (see text).\newline
\textsuperscript{d} 3.0T T2 relaxation time is from dermis, other values are from hypodermis.}
\label{sim_prop} 
\centering                                 
\setlength{\tabcolsep}{3pt}
\begin{tabular}{l |r r r r r r}   
Tissue & $\rho$ & T1(1.5T) & T2(1.5T) & T1(3.0T) & T2(3.0T) & Ref \\
	\midrule
CSF						& 100 (0) & 4326 (0) & 791 (127) & 4313 (0) & 503 (64) & \cite{kroeker1986analysis,cheng1994vivo,rooney2007magnetic,piechnik2009functional}	\\
GM 						& 86 (.4) & 1124 (24) & 95 (8) & 1820 (114) & 99 (7) & \cite{stanisz2005t1}	\\
WM 						& 77 (3) & 884 (50) & 72 (4) & 1084 (45) & 69 (3) & \cite{stanisz2005t1}	\\
Fat 						& 100 (0) & 343 (37) & 58 (4) & 382 (13) & 68 (4) &  \cite{de2004mr}	\\
Muscle 					& 100 (0) & 629 (50) & 44 (6) & 832 (62) & 50 (4) & \cite{stanisz2005t1, barral2009skin}	\\
Skin\textsuperscript{d}		& 100 (0) & 230 (8) & 35 (4) & 306 (18) & 22 (0) & \cite{richard1991vivo, song1997vivo, barral2009skin} \\
Skull\textsuperscript{b}		& 0 (0) & 200 (0) & .46 (0) & 223 (11) & .39 (.02) & \cite{reichert2005magnetic, du2010qualitative}\\
Glial\textsuperscript{a}		& 86 (0) & 1124 (24) & 95 (8) & 1820 (114) & 99 (7) & \cite{cheng1994vivo,stanisz2005t1}	\\
Conn. \textsuperscript{c}	& 77 (0) & 1124 (24) & 95 (8) & 1820 (114) & 99 (7) & \cite{stanisz2005t1}	
\end{tabular} 
\end{table}

\section{$L^p$-norm minimization} \label{app:norms}
In Section \ref{sec:siam_loss} we specified the Siamese loss as the networks objective function. The input of this loss consists of a pairwise distance, for which we chose an $L^{1}$-norm. There are 2 reasons for this: the first is that $L^p$-norms with larger values for $p$ concentrate densely in high-dimensional spaces \cite{flexer2015choosing}. Concentration means that the differences between pairwise distances of a set of points become smaller as the number of dimensions increases. This is a problem because the actions of pulling and pushing will not sufficiently decrease the distance between similar pairs or sufficiently increase the distance between dissimilar pairs. The second reason is that the gradient of the $L^1$-norm is constant, while the gradient of an $L^p$-norms with $p>1$ are functions of the distance \cite{boyd2004convex}. Gradients of norms with large $p$'s become smaller as the distance between pairs becomes smaller, which means the incentive for the network to pull pairs closer decreases. A constant gradient ensures that there will also be a constant incentive to pull similar pairs closer together. Considering that we want our representation to be truly invariant, we want the network to continue to pull similar pairs together until they are as close as possible.

\end{appendices}

\section*{References}

\bibliography{kouw_mia17a}

\end{document}